\documentclass[twoside,11pt,preprint]{article} % for arxiv
\usepackage{jmlr2e_preprint}

\usepackage{color}
\usepackage{multirow}
\usepackage{booktabs}
\usepackage[utf8]{inputenc}
\usepackage[T1]{fontenc}
\usepackage{amsmath,amssymb}
\usepackage{subcaption}
\usepackage{wrapfig}
\usepackage[T1]{fontenc}
\usepackage{pifont}

\newcommand{\cC}{\mathcal{C}}
\newcommand{\covf}{\texttt{coverforest}}
\newcommand{\skl}{\texttt{scikit-learn}}
\newcommand{\cmark}{\ding{51}}%
\newcommand{\xmark}{\ding{55}}%

\usepackage{pythonhighlight}
\ShortHeadings{}{}
\firstpageno{1}

\begin{document}

\title{\covf: Conformal Predictions with Random Forest in Python}

\author{\name Panisara Meehinkong\footnotemark[2]  \\
       \name Donlapark Ponnoprat\footnotemark[2] \footnotemark[1]\phantom{\thanks{Corresponding author.}} \email donlapark.p@cmu.ac.th \\
       \AND
       \addr \footnotemark[2] Department of Statistics,
       Chiang Mai University,
       Chiang Mai 50200, Thailand
       }

%\editor{}

\maketitle

\begin{abstract}%   <- trailing '%' for backward compatibility of .sty file

Conformal prediction provides a framework for uncertainty quantification, specifically in the forms of prediction intervals and sets with distribution-free guaranteed coverage. While recent cross-conformal techniques such as CV+ and Jackknife+-after-bootstrap achieve better data efficiency than traditional split conformal methods, they incur substantial computational costs due to required pairwise comparisons between training and test samples' out-of-bag scores. Observing that these methods naturally extend from ensemble models, particularly random forests, we leverage existing optimized random forest implementations to enable efficient cross-conformal predictions. 

We present \covf, a Python package that implements efficient conformal prediction methods specifically optimized for random forests. Our package leverages parallel computing and \texttt{Cython} optimizations to accelerate out-of-bag calculations, making it 2--10 times faster than existing implementations while maintaining theoretical coverage guarantees. \covf \  supports both regression and classification tasks through various conformal prediction methods, including split conformal, CV+, and Jackknife+-after-bootstrap, as well as adaptive prediction sets for classification. Built upon \skl's random forest implementation, \covf \  provides a practical solution for reliable uncertainty quantification. We demonstrate the package's performance through empirical evaluations on several benchmark datasets. The source code for the \covf \ is hosted on GitHub at \url{https://github.com/donlap/coverforest}.
\end{abstract}


% General comments
% Add motivating example somewhere?
% Models in more detail?

\section{Introduction}

Classification and regression are fundamental machine learning tasks with applications ranging from medical diagnosis to customer segmentation and computer vision. While numerous classification algorithms exist in the literature, most implementations focus solely on point predictions. However, in many real-world applications, particularly those involving critical decision-making, understanding the uncertainty associated with these predictions becomes crucial. For this reason, conformal prediction techniques~\cite{vovk1999,vovk2003,Lei2018,Vovk2022} have recently garnered significant interest, as they leverage data exchangeability to produce prediction sets/intervals that contain the true output with a guaranteed probability above a specified level.

Recent conformal prediction methods like CV+~\cite{Romano2020} and J+ab~\cite{kim2020} rely on out-of-fold and out-of-bag calculations, techniques frequently used in ensemble learning and particularly in random forests \cite{breiman2001}. Years of development in open-source machine learning packages have led to highly optimized out-of-bag calculations in random forests. The \skl's implementation, for instance, leverages parallel computing to speed up the calculations. Even though these implementations are only for in-sample out-of-bag calculations, we can adapt the technique for train-test cross-conformity score calculations, enabling faster conformal predictions with CV+ and J+ab.

Building on this idea, we introduce \covf, a Python package for conformal predictions using random forest. \covf \ implements several recent advances in conformal prediction, including split conformal~\cite{Vovk2022}, CV+~\cite{Barber2021} and J+ab~\cite{kim2020} for regression, and adaptive prediction set (APS)~\cite{Romano2020} with split conformal, CV+, and J+ab for classification. Random forest is our model of choice for three reasons 1) they have consistently demonstrated good performance on tabular data \cite{Grinsztajn2022}, 2) they can be trained with missing data~\cite{breiman2001,Fei2017}, and 3) trees are very fast to train in parallel~\cite{Mitchell2011}, making them attractive for cross-conformal predictions like CV+ and J+ab.

In our implementation, trees are fitted and queried for predictions in parallel using the \texttt{joblib} \cite{joblib} package in Python, and the cross-conformity scores are calculated in \texttt{Cython} \cite{cython}.

The following is a summary of \texttt{coverforest}'s contribution:
\begin{enumerate}
    \item \texttt{coverforest} directly extends \texttt{scikit-learn}'s random forest, which means it also utilizes already well-optimized \texttt{C+} code and parallel computation. Experimental results show that \covf's conformal training and prediction of random forests are 2--10 times faster than an existing implementation.
    \item \texttt{coverforest} is designed to follow \texttt{scikit-learn}'s API conventions with the simple \texttt{fit()} and \texttt{predict()} methods. It can also be easily combined with \texttt{scikit-learn}'s preprocessing methods, allowing it to handle various data formats.
    \item Even though there is an existing package that implements adaptive prediction sets (APS) for conformal prediction, \texttt{coverforest} is the first package to implement the Jackknife+-after-Bootstrap (J+ab) method for conformal prediction.
\end{enumerate}

\section{Background on Conformal Predictions}\label{sec:background}

 Let $(X_1,Y_1),\ldots,(X_n,Y_n)$ be training data for a predictive model $\hat{f}$ and $(X_{n+1},Y_{n+1})$ is a new test point, where $X_i$'s lie in an input space $\mathcal{X}$ and $Y_i$'s lie in an output space $\mathcal{Y}$. Specifically, we let $\mathcal{Y} = \mathbb{R}$ for a regression task and $\mathcal{Y} = \{1,\ldots,C \}$ for a classification task. A \emph{prediction set} for $Y_{n+1}$ is a set of outputs $\cC = \cC(X_i) \subseteq \mathcal{Y}$ that might be input-dependent. For regression tasks we restrict prediction sets to intervals, in which case they are commonly referred to as \emph{prediction intervals}.
 
 In a \emph{conformal prediction} task, we specify a miscoverage rate $\alpha \in (0,1)$, with which we aim to construct a prediction set $\cC$ that has probability at least $1-\alpha$ to contain the true label:
 \begin{equation} \label{eq:2}
  P(Y_{n+1} \in \cC(X_{n+1})) \geq 1-\alpha.
 \end{equation}

Classical results in conformal prediction literature typically assume that the random variables $(X_1,Y_1),\ldots,(X_n,Y_n)$ and $(X_{n+1},Y_{n+1})$ are either i.i.d. or exchangable, that is, their joint distribution are invariant under permutations. Below, we provide a review of well-known conformal prediction methods, and recently proposed conformity that achieve coverage guarantees under one of these assumptions.

\subsection*{Regression}
\begin{enumerate}
\item \textbf{Split Conformal \cite{Vovk2022}.} First, we split the data into training and calibration sets, we fit a predictive model $\hat{f}$ on the training data. The method then computes nonconformity scores $R_i = |Y_i - \hat{f}(X_i)|$ for each point in the calibration set. For the new test point $X_{n+1}$, the prediction interval is constructed as: 
\[ \cC(X_{n+1}) = [\hat{f}(X_{n+1}) - q_{\alpha}, \hat{f}(X_{n+1}) + q_{\alpha}], \] 
where $q_{\alpha}$ is the $\lceil(1-\alpha)(n+1)\rceil$-th smallest value of the calibration scores.

\item \textbf{Cross-Validation+ (CV+) \cite{Barber2021}.} Split the data into $K$ folds. For each fold $\mathcal{I}_k$, we fit a model $\hat{f}^{(-k)}$ on all data except fold $k$ and compute residuals $R_i = |Y_i - \hat{f}^{(-k)}(X_i)|$ for all $i \in \mathcal{I}_k$. The final prediction interval for a new point is constructed as:
\[ \cC(X_{n+1}) = [q^-_{\alpha}\{\hat{f}^{-S(i)}(X_{n+1}) - R_i\}, q^+_{\alpha}\{\hat{f}^{-S(i)}(X_{n+1}) + R_i\}],\] 
where $S(i) = k$ if $i \in \mathcal{I}_k$, $q^+_{\alpha}\{a_i\}$ is the $\lceil(1-\alpha)(n+1)\rceil$-th smallest value of $a_i$'s and $q^-_{\alpha}\{a_i\}$ is the $\lfloor \alpha(n+1)\rfloor$-th smallest value of $a_i$'s.

\item \textbf{Jackknife+-after-Bootstrap (J+ab) \cite{kim2020}.} This method is similar to CV+ but instead of splitting data into $K$-folds, we sample with or without replacement from the original training set multiple times, and fit a model on each of the subsamples. During prediction, we let $\hat{f}^{(-i)}(X_i)$ be the aggregated predictions from all models that were fitted on the subsamples that do not contain $(X_i,Y_i)$, from which we compute the residuals $R_i = |Y_i - \hat{f}^{(-i)}(X_i)|$. The final prediction interval for a new point is then 
\begin{equation} \label{eq:jab} 
\cC(X_{n+1}) = [q^-_{\alpha}\{\hat{f}^{(-i)}(X_{n+1}) - R_i\}, q^+_{\alpha}\{\hat{f}^{(-i)}(X_{n+1}) + R_i\}],
\end{equation} 
where $q^-_{\alpha}$ and $q^+_{\alpha}$ are defined as in CV+. However, this interval with a fixed number of bootstrap samplings $B$ does not achieve a finite-sample coverage guarantee, due to the asymmetry between training and test samples; to see this, observe that each test sample is excluded from exactly $B$ bootstraps, while each training sample is typically excluded from $<B$ bootstraps. This issue is resolved by resampling $B$ from a Binomial distribution: 
\begin{equation}\label{eq:binom} 
\operatorname{Binomial}(\tilde{B}, (1- \frac1{n+1})^m),
\end{equation}
where $\tilde{B}$ is a fixed initial value, $n$ is the number of training samples and $m$ is the number of bootstrapped samples. One can show that the interval in \eqref{eq:jab} with the randomized $B$ restores the symmetry between training and test samples \cite{kim2020}.
\end{enumerate}

\subsection*{Classification}

\begin{enumerate}
\item \textbf{Split Conformal with APS Score \cite{Romano2020}.} We split the data into training and calibration sets, then fit a classification model $\hat{f}$ on the training set. An obvious conformity scores are the model's probability predictions; however, a hypothetical model that assigns $50-50$ probabilities to the most likely and second most likely classes would result in set predictions that only contain the most likely classes, and hence fails to achieve the target coverage level. Alternatively, Adaptive Prediction Sets (APS) \cite{Romano2020} has been recently proposed method that is adaptive to data distributions.

The crux of the APS method is the use of cumulative probability as the conformity score: given the predicted probabilities of a point $(x,y)$ in descending order $\hat{\pi}_{1}(x) \geq \hat{\pi}_{2}(x)\geq\ldots \geq \hat{\pi}_{C}(x)$, the \emph{APS score function} is the (randomized) \emph{generalized inverse quantile}: 
\begin{equation} \label{eq:APSscore} E(x,y, \hat{\pi}) = \hat{\pi}_{1}(x) + \ldots + \hat{\pi}_{y-1}(x) + u\cdot \hat{\pi}_y(x), \quad u \sim \operatorname{Uniform}(0,1).
\end{equation}
In other words, the score is chosen randomly between the cumulative probabilities of $y$ and $y-1$. Without this randomization, the prediction sets would generally cover more than the desired level of ($1-\alpha$)\% if $E(x,y, \hat{\pi})$ is defined up to $\hat{\pi}_y$, and less if it is defined up to $\hat{\pi}_{y-1}$.

In the case of split conformal prediction, we let $q_{\alpha}$ be the $\lceil(1-\alpha)(n+1)\rceil$-th smallest value of the calibration scores. The prediction set for a new point $X_{n+1}$ is then: 
\[ \cC(X_{n+1}) = \left\{1,\ldots, k :k = \inf_{y \in \{1,\ldots, C\}} E(X_{n+1}, y, \hat{\pi}) \geq q_{\alpha}\right\}.\] 

The APS score typically produces larger prediction sets than the probability prediction score; however, the former generally achieves the target coverage level while the latter often fails to do so. In Section~\ref{sec:experiments}, we experimentally compare these two scores to confirm this observation.

\item \textbf{Cross-Validation+ (CV+) with APS Score \cite{Romano2020}.} The CV+ method for classification aims to improve data efficiency compared to a simple train-calibration split by using a cross-validation scheme. This allows each data point to be used for both training and calibration, but never in the same role simultaneously. The procedure using the Adaptive Prediction Set (APS) score is as follows:

\begin{enumerate}
    \item \textbf{Data Partitioning.} Split the training data $\{(X_i, Y_i)\}_{i=1}^n$ into $K$ disjoint folds, denoted by $\mathcal{I}_1, \ldots, \mathcal{I}_K$.
    
    \item \textbf{Cross-Validation Training.} For each fold $k \in \{1, \ldots, K\}$, train a classification model $\hat{\pi}^{(-k)}$ on all data points \textit{except} those in fold $\mathcal{I}_k$. This results in $K$ different models.
    
    \item \textbf{Leave-one-fold-out Score Calculation.} We associate each data point $(X_i, Y_i)$, with the model $\hat{\pi}^{(-S(i))}$ where $S(i) = k$ if $i \in \mathcal{I}_k$ (i.e., the model that was not trained on the fold containing $i$) to calculate its APS conformity score. Given the predicted probabilities for $(X_i, Y_i)$ sorted in descending order $\hat{\pi}_{1}^{(-S(i))}(X_i) \geq \ldots \geq \hat{\pi}_{C}^{(-S(i))}(X_i)$, the APS score is
    $E_i = E(X_i,Y_i, \hat\pi^{-S(i)})$, where the generalized inverse quartile $E$ is defined in \eqref{eq:APSscore}. This process yields a set of $n$ conformity scores $\{E_1, \ldots, E_n\}$.
    
    \item \textbf{Prediction Set Construction.} For a new test point $X_{n+1}$ and a desired miscoverage rate $\alpha$, we construct the prediction set $\mathcal{C}(X_{n+1})$ as follows: With the conformity scores $E_1,\ldots,E_n$ obtained from the previous step, the $p$-\emph{value} for a candidate class $y$ is given by:
    \begin{equation} \label{eq:pval_cv}
    \hat{p}(y) = \frac{1}{n} \sum_{i=1}^n \mathbb{I}\left\{ E_i \geq E(X_{n+1}, y, \hat{\pi}^{(-S(i))}) \right\}.
    \end{equation}
    The prediction set is then formed by including all classes whose $p$-value is greater than or equal to $\alpha$:
    \begin{equation} \label{eq:pred_set_cv}
    \mathcal{C}(X_{n+1}) = \{ y \in \{1, \ldots, C\} \mid \hat{p}(y) \geq \alpha \}.
    \end{equation}
\end{enumerate}

\item \textbf{Jackknife+-after-Bootstrap (J+ab) with APS Score \cite{kim2020}.} The J+ab method is similar to CV+ but utilizes bootstrapping instead of $K$-fold cross-validation to create leave-one-out models and scores. This approach is particularly well-suited for ensemble methods like random forests.

\begin{enumerate}
    \item \textbf{Bootstrap Ensemble.} We generate $B$ bootstrap samples from the training data $\{(X_i, Y_i)\}_{i=1}^n$. For each bootstrap sample $b \in \{1, \ldots, B\}$, train a classification model $\hat{\pi}^{(b)}$.

    As with the J+ab for regression, to ensure a valid finite-sample coverage guarantee, the number of bootstrap samples, $B$, must be randomized according to \eqref{eq:binom}.
    
    \item \textbf{Leave-one-out Predictions and Scores.} For each training point $(X_i, Y_i)$, we create an aggregated leave-one-out predictor, $\hat{\pi}^{(-i)}$, by averaging the predictions of all models $\hat{\pi}^{(b)}$ that were trained on bootstrap samples that did \textit{not} include the point $(X_i, Y_i)$. Using this leave-one-out predictor, calculate the APS score
    $E_i = E(X_i,Y_i, \hat\pi^{(-i)})$, where the generalized inverse quartile $E$ is defined in \eqref{eq:APSscore}.
    
    \item \textbf{Prediction Set Construction.} For a new test point $X_{n+1}$, the prediction set is constructed by calculating a $p$-value using \eqref{eq:pval_cv}, but with $\hat{\pi}^{(-i)}$ replacing $\hat{\pi}^{(-S(i))}$:
    \[ \hat{p}(y) = \frac{1}{n} \sum_{i=1}^n \mathbb{I}\left\{ E_i \geq E(X_{n+1}, y, \hat{\pi}^{(-i)}) \right\}. \]
    The final prediction set is then formed by including all classes with a sufficiently high $p$-value, as in \eqref{eq:pred_set_cv}:
    \[ \mathcal{C}(X_{n+1}) = \{ y \in \{1, \ldots, C\} \mid \hat{p}(y) \geq \alpha \}. \]
\end{enumerate}
\end{enumerate}

\emph{Regularized Adaptive Prediction Sets (RAPS)} \cite{angelopoulos2021un}. This score was proposed for classification with a lot of classes (say 1000+ classes) where the ordering of the predicted probabilities does not reflect the true order. To this end, the authors propose to add to the APS's score function a regularization term that penalizes classes with small predicted probabilities (assuming $\hat{\pi}_{1}(x) \geq \hat{\pi}_{2}(x)\geq\ldots \geq \hat{\pi}_{C}(x)$ as before):
\begin{equation} \label{eq:RAPSscore}
    E_{k,\lambda}(x,y) = E(x,y) + \lambda \cdot  \max(0, y - k).
\end{equation}
As with APS, we can use this score function for split conformal, CV+, or J+ab.

\begin{table}[t]
    \centering 
    \begin{tabular}{|l|c|c|c|c|}
        \hline
        \multirow{2}{*}{\textbf{Method}} & \textbf{Theoretical} & \textbf{Practical} & \textbf{Training} & \textbf{Prediction} \\
        & \textbf{Coverage} & \textbf{Coverage} & \textbf{Cost} & \textbf{Cost} \\
        \hline
        \multicolumn{5}{|l|}{\textbf{Regression}} \\
        \hline
        Split Conformal & $1-\alpha$ & $\approx 1-\alpha$ & $O(T dn\log n)$ & $O(T n_{\operatorname{test}} \log n)$ \\
        \hline
        CV+ & $1-2\alpha-\epsilon_{K,n}$ & $\gtrsim 1-\alpha$ & $O(KTdn \log n)$ & $O(KT n_{\operatorname{test}} \log n)$ \\
        \hline
        J+ab & $1-2\alpha$ & $\gtrsim 1-\alpha$ & $O(Tdn\log n)$ & $O(n * n_{\operatorname{test}})$ \\
        \hline
        \multicolumn{5}{|l|}{\textbf{Classification}} \\
        \hline
        Split + (R)APS & $1-\alpha$ & $\approx 1-\alpha$ & $O(Td n\log n)$ & $O(T  n_{\operatorname{test}} \log n)$ \\
        \hline
        CV+ + (R)APS& $1-2\alpha-\epsilon_{K,n}$ & $\gtrsim 1-\alpha$ & $O(KTdn\log n)$ & $O( Cn * n_{\operatorname{test}})$ \\
        \hline
        J+ab + (R)APS & $1-2\alpha$ & $\gtrsim 1-\alpha$ & $O(Tdn\log n)$ & $O(Cn * n_{\operatorname{test}})$ \\
        \hline 
    \end{tabular}
    \caption{Comparison of coverage guarantees and computational costs for different conformal random forest methods. Here, the notation $\gtrsim 1-\alpha$ means ``sometimes worse than $1-\alpha$'', $n$ is the number of training samples, $n_{\operatorname{test}}$ is the number of test samples, $T$ is the number of trees, $C$ is the number of classes, $K$ is the number of folds in CV+, and $\epsilon_{K,n} = \min \left\{ 2(1 - 1/K)/(n/K + 1), (1-K/n)/(K+1) \right\}$. We assume that  $T \ll n$ and $C \le T$.}
    \label{tab:conformal_comparison}
\end{table}

In Table~\ref{tab:conformal_comparison}, we compare coverage guarantees and computational costs across different conformal prediction methods for random forests. While CV+ and J+ab theoretically guarantee coverages of $\approx 1-2\alpha$, they typically achieve $\geq 1-\alpha$ coverage in practice. One downside of CV+ and J+ab with the APS and RAPS scores is the notably long prediction times due to the pairwise comparisons between train and test conformity scores across all classes, which contribute to $Cn * n_{\text{test}}$ computational cost.

\subsection{Related work}
The \covf \ package implements conformal predictions \cite{Lei2018,Vovk2022, Vovk2009,vovk1999,vovk2003} using random forest as the underlying model \cite{breiman2001}. For similar work, conformity scores of bootstrapped regression based on iterative reweighted least squares is briefly discussed in \cite{Vovk2022}. For random forest classification, two conformity scores based on out-of-bag predictions were proposed in~\cite{Dmitry2010}. However, a finite-sample coverage guarantee cannot be obtained as they do not preserve exchangability~\cite{kim2020}.

Nonconformity scores based on out-of-bag (OOB) predictions were introduced in \cite{Johansson2014}. The experiments revealed that these scores yield smaller prediction intervals compared to neural networks and $k$-nearest neighbors. Later, it is confirmed in \cite{Zhang2020} that these intervals provide asymptotic coverage guarantee, which is only attained with infinitely many data. Further works explored various aspects of out-of-bag calibration for conformal predictions \cite{Bostrom2017, Lofstrom2013, Linusson2020}. \covf's implementation of Jackknife+-after-bootstrap uses a randomized number of bootstraps in order to achieve a finite-sample coverage guarantee, which is proved in~\cite{kim2020}.

A conformal regression method related to CV+ is Jackknife+ \cite{Barber2021} where the regression model is fitted on leave-one-out datasets. It is a special case of CV+ where each fold only consists of a single training point, and the number of folds equals the number of training samples. \covf \ can be used with the Jackknife+ method by specifying \texttt{method="cv"} and \texttt{cv} equals the number of training samples. Another method is the conformalized quantile regression (CQR) which constructs prediction intervals based on classical quantile regression. Implementing CQR is one of our future plans where we can incorporate existing implementation of conformal quantile random forest \cite{Meinshausen06} such as \texttt{quantile-forest} \cite{Johnson2024}.

For classification, given a model that outputs probability estimates $\hat{\pi}_j(x) \approx P(Y=j \vert X=x)$, a basic conformal prediction method can be devised based on a score function $s(x,y) = 1 - \hat{\pi}_y(x)$. The prediction set for a new point $X_{n+1}$ is then given by $\cC(X_{n+1}) = \left\{ y : s(X_{n+1}, y) \leq q_{\alpha}\right\}$, where $q_{\alpha}$ is the $\lceil(1-\alpha)(n+1)\rceil$-th smallest value of the calibration scores \cite{Sadinle2019,Vovk2022}. However, this method tend to outputs an empty set when the model is uncertain about the prediction, for example, $q_{\alpha} =0.2$ and the predicted probabilities of two classes are 0.5 each. This issue can be remedied by instead using the cumulative probability as in the APS method \cite{Romano2020}. Our future plans include implementing multilabel classification with finite-sample coverage guarantee via conformal risk control \cite{Bates2021,angelopoulos2021a,angelopoulos2024}.

There are several software packages that implement various types of conformal predictions, such as inductive conformal prediction and mondrian conformal prediction. However, \texttt{coverforest} is the first package to implement CV+, J+ab and split conformal with both APS and RAPS scores for set predictions tasks. See Table~\ref{tab:conformal_comparison} for a feature comparison between \texttt{coverforest} and other conformal prediction packages. We will also see in Section~\ref{sec:experiments} that \texttt{coverforest}'s implementations of APS and RAPS-based set predictions are faster than the alternative (\texttt{MAPIE}). by 2 to 5 times

\begin{table}[] \footnotesize
\begin{tabular}{@{\extracolsep{4pt}}lccccccc@{}}
\textbf{}          & \multicolumn{2}{c}{\textbf{Score}} & \multicolumn{5}{c}{\textbf{Method}}        \\ \cmidrule{2-3} \cmidrule{4-8} 
\textbf{Package} & \textbf{APS} & \textbf{RAPS} & \textbf{CV+} & \textbf{J+ab} & \textbf{Split} & \textbf{Inductive} & \textbf{Mondrian} \\ \midrule
nonconformist~\cite{linusson2023nonconformist}      & \xmark           & \xmark          & \xmark & \xmark & \cmark & \cmark & \xmark \\
Orange3-Conformal~\cite{Hocevar2021}  & \xmark           & \xmark          & \xmark & \xmark & \cmark & \cmark & \xmark \\
conforest~\cite{Johansson2014}          & \xmark           & \xmark          & \xmark & \xmark & \xmark & \cmark & \xmark \\
crepes~\cite{bostrom2024}      & \xmark           & \xmark          & \xmark & \xmark & \cmark & \cmark & \cmark \\
\texttt{MAPIE}~\cite{Cordier2023}   & \cmark           & \cmark          & \cmark & \xmark & \cmark & \xmark & \xmark \\
coverforest        & \cmark           & \cmark          & \cmark & \cmark & \cmark & \xmark & \xmark \\ \bottomrule
\end{tabular}
\label{tbl:comparepackages}
\caption{Feature comparison between various conformal prediction software packages for set prediction tasks.}
\end{table}

\section{Implementation}

\subsection{API description}
\begin{figure}[h]
  \centering
    \includegraphics[width=\textwidth]{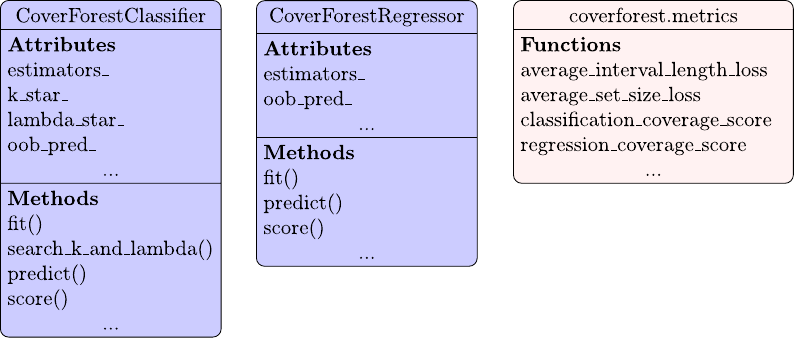}
  \caption{Class structure and the metrics in \covf.}
  \label{fig:api}
\end{figure}

\covf's provides two main model classes with familiar \skl-style API, with the standard \texttt{fit()} and \texttt{predict()} methods, as shown in Figure~\ref{fig:api}. The most important parameters of both classes are:
\begin{itemize}
\item \texttt{n\_estimators}: The number of sub-estimators in the forest
\item \texttt{method}: The conformal prediction methods, which can be one of \texttt{cv} (CV+), \texttt{bootstrap} (Jackknife+-after-bootstrap), and \texttt{split}, (split conformal)
\item \texttt{cv}: The number of folds in the CV+ method
\end{itemize}
There are also several parameters that are only specific to APS and RAPS: 
\begin{itemize}
\item \texttt{k\_init\_} and \texttt{lambda\_init\_}: The regularization parameters in RAPS. The APS score is equivalent to setting \texttt{k\_init\_=0} and \texttt{lambda\_init\_=0}. Sutomatic searching using the method described in \cite{angelopoulos2021un} is also available by setting \texttt{k\_init\_="auto"} and  \texttt{lambda\_init\_="auto"}
\item \texttt{allow\_empty\_set}: whether to allow empty set predictions
\item \texttt{randomized}: whether to apply the APS's randomization scheme to include an additional class in prediction sets as described in Section~\ref{sec:background}.
\end{itemize}
There are only one parameter that is specific to the Jackknife+-after-bootstrap method: 
\begin{itemize}
\item \texttt{resample\_n\_estimators}: whether to draw the number of trees from the Binomial distribution \eqref{eq:binom}, in which case \texttt{n\_estimators} is taken as the first parameter $\tilde{B}$ of the distribution.
\end{itemize}

We now go over the methods in both model classes.

The \texttt{fit()} method fits each tree on the training data. The trees are fitted in parallel via a helper class \texttt{joblib.Parallel()} \cite{joblib}. The fitted trees are then used to calculate for each training sample the out-of-box (OOB) predictions, which are then used to calculate the calibration scores. During fitting, the method maintains $\mathcal{T}_i$ for each sample $(X_i, Y_i)$, where $\mathcal{T}_i$ is the collection of trees not fitted on this particular sample.

The \texttt{predict()} method first queries predictions from all trees for any given test sample $(X_{n+1}, Y_{n+1})$, where the queries are performed in parallel via \texttt{joblib.Parallel()}. Subsequently, for each training instance $i$, it aggregates predictions exclusively from trees in $\mathcal{T}_i$, converting these into a conformity score. This score is then compared with the calibration scores of $(X_i, Y_i)$ to produce the final predictions.

For moderately large training and test sets, calculating the conformity scores across all pairs of training and test samples can be quite expensive. We contribute to performance improvement by implementing cross-calculations in \texttt{Cython} with parallel processing enabled. Specifically, the \texttt{\_giqs.pyx} script in the package contains optimized code that speeds up calculation of cross-conformity scores, resulting in substantially faster prediction time than the previous package. Users can specify the number of threads through the \texttt{num\_threads} parameter in the \texttt{predict()} method.

The \texttt{CoverForestClassifier} includes the \texttt{search\_k\_and\_lambda()} method for automatic parameter optimization. This method determines optimal values for $k$ and $\lambda$ according to the approach detailed in \cite{angelopoulos2021un}.

\covf \ also has a \texttt{metric} module that provides convenient functions for evaluating the prediction sets/intervals by measuring their average size and empirical coverage (e.g. the proportion of prediction sets/intervals that contains the true class/value).

\subsection{Computational complexity and memory usage}

As summarized in Table~\ref{tab:conformal_comparison}, both CV+ and J+ab methods require $C n * n_{\text{test}}$ memory and computations for the calculation of conformity scores. To reduce the memory, the user could split the test set into several chunks and make predictions on each chunk. Alternatively, the user could use the split conformal method which requires only $n * n_{\text{classes}}$ memory and $T n_{\text{test}} \log n$ computations while being able to achieve satisfactorily small set sizes on a large dataset.

\subsection{Source code}

The source code for the \covf \ is hosted on GitHub at \url{https://github.com/donlap/coverforest}. The code follows the PEP8 guidelines, and all docstrings of publicly exposed classes comply with the NumPy docstring style. Both \texttt{CoverForestClassifier} and \texttt{CoverForestRegressor} pass the comprehensive unit tests provided by \texttt{scikit-learn}. The documentation includes installation guides, user manual, and API reference, generated using sphinx and available online at \url{https://donlapark.github.io/coverforest}. The package is BSD-licensed and requires only \texttt{scikit-learn} version 1.6.1 or greater.

\section{Examples}

We demonstrate how to use \covf \ on a toy dataset for set predictions.
\subsection*{Conformal predictions with CV+ and RAPS score} 
We train a conformal random forest using CV+ with the RAPS score. To use the CV+ method, initialize a \texttt{CoverForestClassifier} model with \texttt{method="cv"}. Here, we set \texttt{cv=5} to split the data into five folds. As a result, five random forests will be fitted on the five combinations of 4-folds. We also specify that each forest has 200 trees by setting \texttt{n\_estimators=200}. 
\begin{python}
from coverforest import CoverForestClassifier
from coverforest.metrics import average_set_size_loss
from coverforest.metrics import classification_coverage_score
from sklearn.datasets import load_digits
from sklearn.model_selection import train_test_split

X, y = load_digits(return_X_y=True)
X_train, X_test, y_train, y_test = train_test_split(X, y, test_size=0.2)

clf = CoverForestClassifier(n_estimators=200, method="cv", cv=5)
\end{python}
\subsection*{Training and making predictions} 
We train the random forest by calling the \texttt{fit()} method, and make a prediction by calling the \texttt{predict()} method. During prediction, we choose the miscoverage rate at $0.05$ by specifying \texttt{alpha=0.05}. The output of the \texttt{predict()} method is a tuple: The first item contains the single class predictions by the random forest, and the second item contains the set predictions.
\begin{python}
clf.fit(X_train, y_train)
y_pred, y_sets = clf.predict(X_test, alpha=0.05)
# y_pred[:5] = [2, 8, 2, 6, 6]
# y_sets[:5] = [array([2]), array([2, 5, 8]), array([2]), 
#               array([5, 6]), array([6])]
\end{python}
\subsection*{Computing average set size and coverage} 
We may use the provided \texttt{metrics} functions to evaluate the set predictions against the true classes.
\begin{python}
avg_set_size = average_set_size_loss(y_test, y_sets)
coverage = classification_coverage_score(y_test, y_sets)
# avg_set_size = 1.589
# coverage = 0.964
\end{python}
The small average set size indicates that the model can make predictions with high confidence. The coverage tells us that $96.4\%$ of the prediction sets contain the true classes, which is above our specified level of $95\%$.

\subsection*{Conformal predictions using Jackknife+-after-bootstrap with the APS score} 
In the second example, we train a conformal random forest using J+ab with the APS score. To use J+ab, initialize a \texttt{CoverForestClassifier} model with \texttt{method="bootstrap"}. To use the APS method, set \texttt{k\_init=0} and \texttt{lambda\_init=0}. In contrast to CV+ which has multiple forests, J+ab has only a single forest. Here, we specify that this forest has 400 trees.
\begin{python}
clf = CoverForestClassifier(
    n_estimators=400, method="bootstrap", k_init=0, lambda_init=0
)
clf.fit(X_train, y_train)
y_pred, y_sets = clf.predict(X_test, alpha=0.05)
avg_set_size = average_set_size_loss(y_test, y_sets)
coverage = classification_coverage_score(y_test, y_sets)
# y_pred[:5] = [2, 8, 2, 6, 6]
# y_sets[:5] = [array([2]), array([2, 5, 8, 9]),
#               array([2]), array([6]), array([6])]
# avg_set_size = 1.761
# coverage = 0.968
\end{python}
Compared to the CV+ method, J+ab produces prediction sets with a larger average with practically the same coverage.

\section{Experiments and Discussions} \label{sec:experiments}

To validate our implementation, we benchmark \covf's three methods: CV+, J+ab, and split conformal on conformal classfication and regression tasks. We compare our implementations against the most recent packages for conformal predictions, namely the \texttt{crepes} and \texttt{MAPIE} packages.

\subsection{Classification experiment}
We evaluate the predictions and run times of \texttt{CoverForestClassifier} on five classification datasets, namely Mice~\cite{Micedata}, WineQuality~\cite{winedata}, Bean~\cite{beandata}, MNIST, and Helena~\cite{Automl2019} datasets. See Table~\ref{tab:exp} for the datasets' characteristics.

We aim to evaluate the methods' abilities to achieve the desired level of coverage even with a few training points. To this end, we randomly select 20\% of the dataset as the test set (except the MNIST dataset whose test set has already been prepared), and 200 rows of the remaining data as the training set. For the split conformal method, we reserve 100 rows of the training set as the calibration set. We then fit the models on the training set with the target miscoverage level $\alpha \in \{0.05, 0.1, 0.2\}$. We repeat the experiment 50 times for each method and compare the empirical coverage (the proportion of prediction sets that contain the true class), and average set size. 

For CV+, we split the training data into 10 folds, and each random forest has 30 trees. For J+ab and split conformal, the random forest has 100 trees.

\begin{table}[t]
\footnotesize
\centering

\begin{tabular}{@{}rrrrrrr@{}}
\toprule
\multicolumn{1}{l}{} &  &  & \multicolumn{2}{c}{\textbf{Experiment 1 \& 2}} & \multicolumn{2}{c}{\textbf{Experiment 3}} \\
\cmidrule(lr){4-5} \cmidrule(lr){6-7}
& \textbf{Classes} & \textbf{Features} & \textbf{Train} & \textbf{Test} & \textbf{Train} & \textbf{Test} \\
\midrule
Mice \cite{Micedata}         & 8       & 80     & 200     & 216   & 861 & 216   \\
WineQuality \cite{winedata}  & 7       & 11     & 200     & 400   & 3918  & 980  \\
Bean \cite{beandata}   & 7             & 16     & 200    & 2723   & 5000 & 1250   \\
MNIST \cite{MNISTdata}       & 10      & 784     & 200    & 10000  & 5000 & 1250     \\  
Helena \cite{Automl2019}       & 100      & 27     & 1033     & 300  & 904 & 300      \\ \bottomrule
\end{tabular}
\caption{The datasets used in the conformal classification experiments. The smaller training set of the Helena dataset in Experiment 3 is due to the memory limit of the \texttt{MAPIE} package.}

\label{tab:exp}
\end{table}

We also compare our methods against \texttt{crepes}~\cite{bostrom2022}, which uses the classifier's predicted probabilities as the conformity scores; however, such scores are not adaptive in the sense that they tend to result in coverages that are over or under the nominal level, depending on the accuracy of the classifier. 

\texttt{crepes} offers the OOB and split conformal methods. In our experiment, we run both methods with a random forest with 100 trees as the base model.

As the \texttt{MAPIE} library implements the same CV+ and split conformal methods as ours, we do not run \texttt{MAPIE} in this experiment. Nonetheless, we will compare our run times against \texttt{MAPIE} in a later experiment. 

For Experiment 1 and 2, we also perform hypothesis testing to verify that the empirical coverage probabilities exceed the nominal coverage levels, and to compare the average set sizes between any pair of the methods. The results and discussion of the tests can be found in \ref{sec:tests}.

\begin{figure}
  \centering
  \includegraphics[width=1\textwidth]{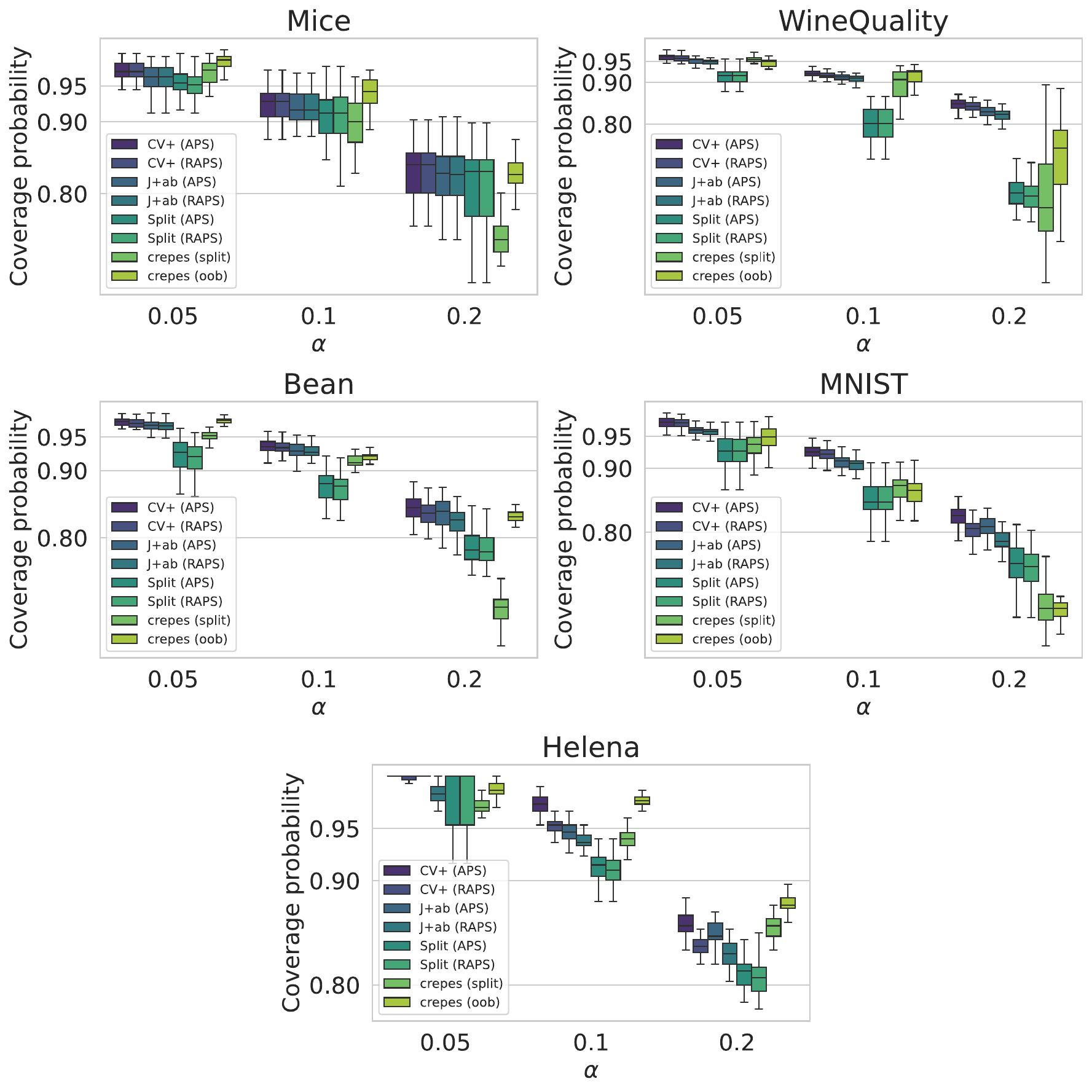} 
  \caption{Coverage probabilities of \texttt{CoverForestClassifier}'s set predictions on the test set, repeated 50 times for each of the CV+, J+ab and split conformal methods with the APS and RAPS scores. The results of \texttt{crepes}'s methods are also included.}
  \label{fig:boxplot1}
\end{figure}

\subsection*{Experiment 1. Empirical coverage}

Figure~\ref{fig:boxplot1} presents the empirical coverages achieved using both the APS \eqref{eq:APSscore} and RAPS \eqref{eq:RAPSscore} scores. The plots consistently demonstrate that CV+ provides the largest coverage. J+ab consistently yields slightly lower coverages than CV+. On the Mice, Bean, and Helena datasets, the OOB method from \texttt{crepes} results in notably large coverage across all values of $\alpha$. In contrast, the split conformal method provides the lowest and most inconsistent coverages on these same datasets.

We now turn our attention to the effect of regularization on the empirical coverages. In this regard, the CV+ and J+ab methods, when paired with the RAPS score, yield slightly reduced coverages for $\alpha=0.1$ and $0.2$. This difference becomes more pronounced in datasets with a large number of classes, as is evident in the results for the MNIST and Helena datasets.

Finally, we assess whether these methods achieve their nominal coverage levels. The plots indicate that the median coverages for both the CV+ and J+ab methods exceed these target levels in the majority of cases. This outcome demonstrates the data efficiency of these two approaches, which allows them to successfully reach the desired coverage.
\subsection*{Experiment 2. Average set size}

\begin{figure}
    \centering
    \includegraphics[width=1\textwidth]{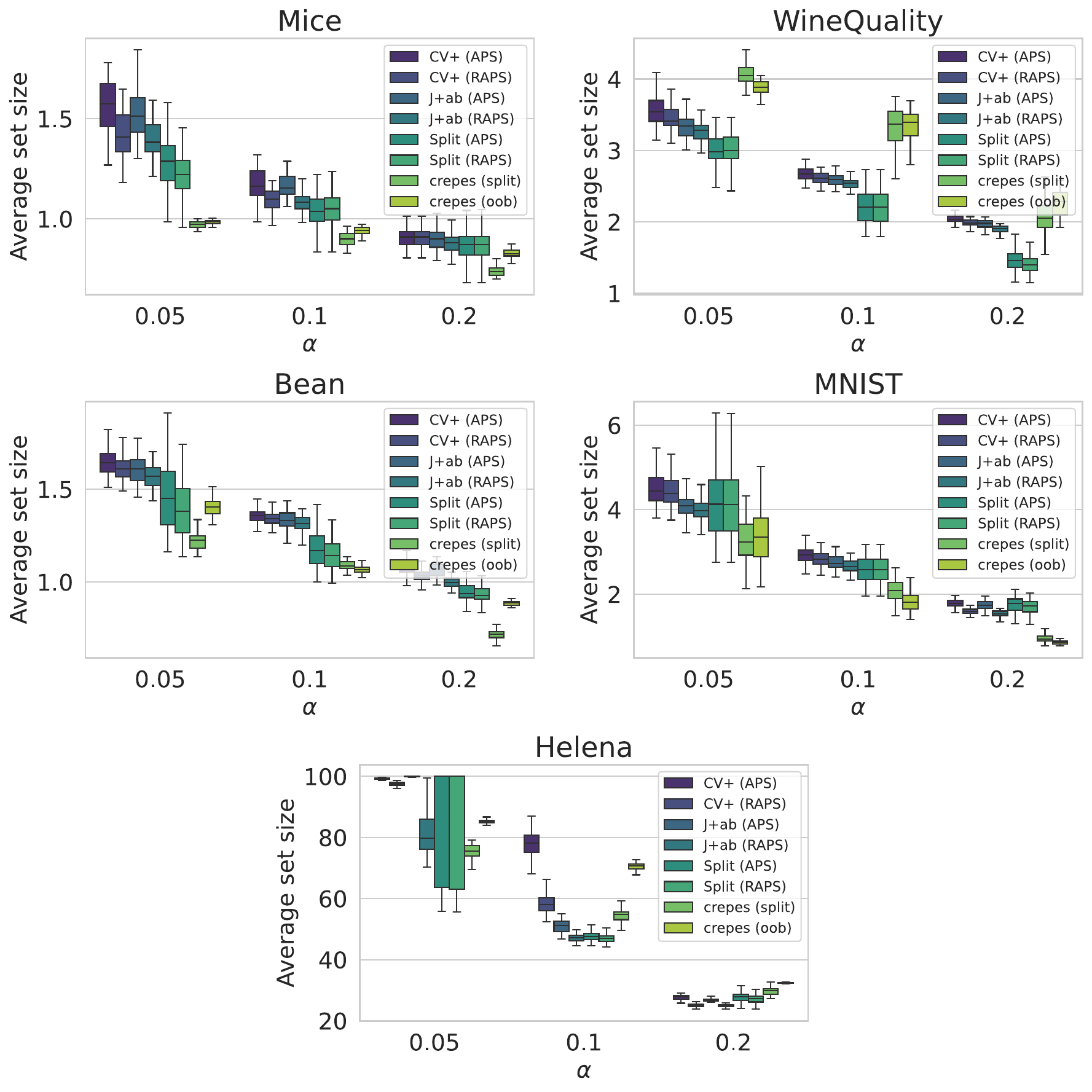}
    \caption{Average size of \texttt{CoverForestClassifier}'s set predictions on the test set, repeated 50 times for each of the CV+, J+ab and split conformal methods with the APS and RAPS scores. The results of \texttt{crepes}'s methods are also included.}
    \label{fig:boxplot2}
\end{figure}

Figure~\ref{fig:boxplot2} illustrates the average prediction set sizes produced by each method. Across all datasets, we observe a general trend where CV+ produces the largest average set sizes, followed by J+ab, and then the split conformal method. The split and OOB methods from \texttt{crepes} yield the smallest set sizes on the Mice, Bean, and MNIST datasets. It is important to recall from the previous experiment, however, that these methods did not achieve the nominal coverage levels on the MNIST dataset.

We now examine the impact of regularization on the average set sizes. For all datasets, the application of the RAPS score contributes to a reduction in the set sizes for the CV+, J+ab, and split conformal methods. This decrease in set size is substantially more noticeable for datasets with a large number of classes, as demonstrated by the results on the Helena dataset.

\subsection*{Experiment 3. Run time comparison with \texttt{MAPIE}}

In this experiment, we compare the run time of \texttt{CoverForestClassifier} against \texttt{MAPIE} \cite{Cordier2023}. As far as we know, it is currently the only package that provides the CV+ methods with the APS and RAPS scores (the J+ab method is also provided in \texttt{MAPIE} but only for regression tasks).

We do not compare our methods against the \texttt{crepes} library here as it does not provide the CV+ and J+ab methods, nor the APS and the RAPS scores.

For \covf \ and \texttt{MAPIE}, we train a random forest and make conformal predictions using CV+, J+ab and split conformal methods, all with the APS score. The numbers of training and test instances in this experiment are indicated in Table~\ref{tab:exp}. We run CV+ with 100 forests with 10 trees each, and J+ab and split conformal with 900 trees. And we fix $\alpha=0.05$. We run each method for 30 repetitions each and report the average training and prediction times. 

\begin{figure}
    \centering
    \includegraphics[width=1\textwidth]{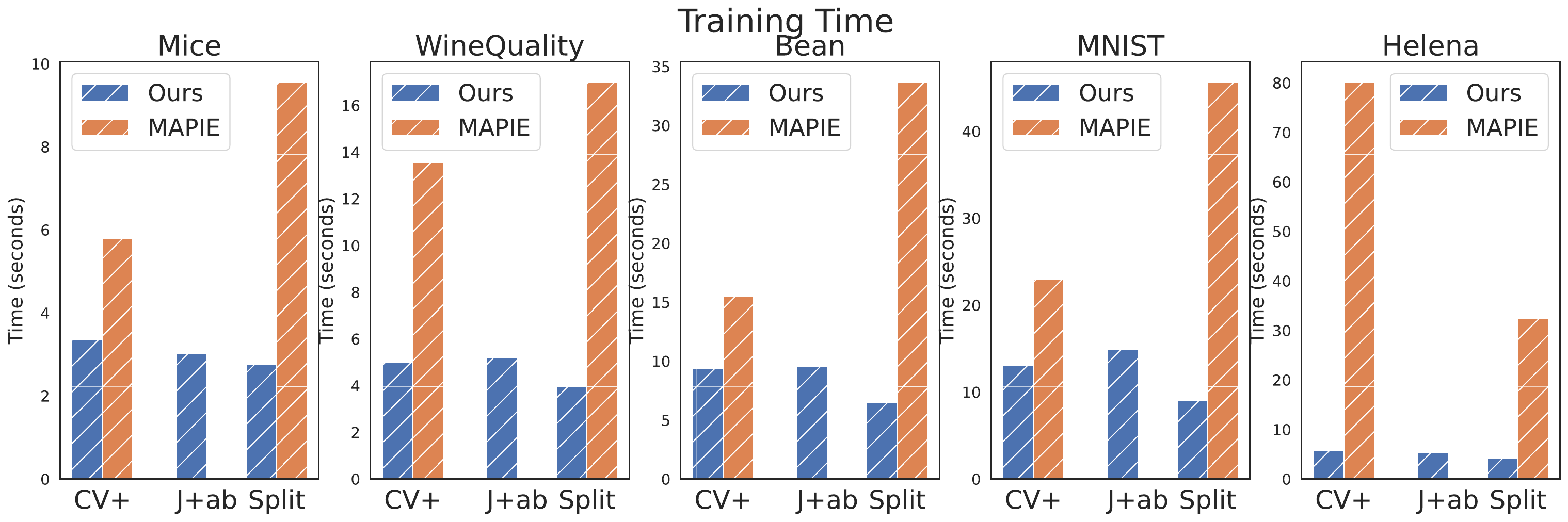}
    \includegraphics[width=1\textwidth]{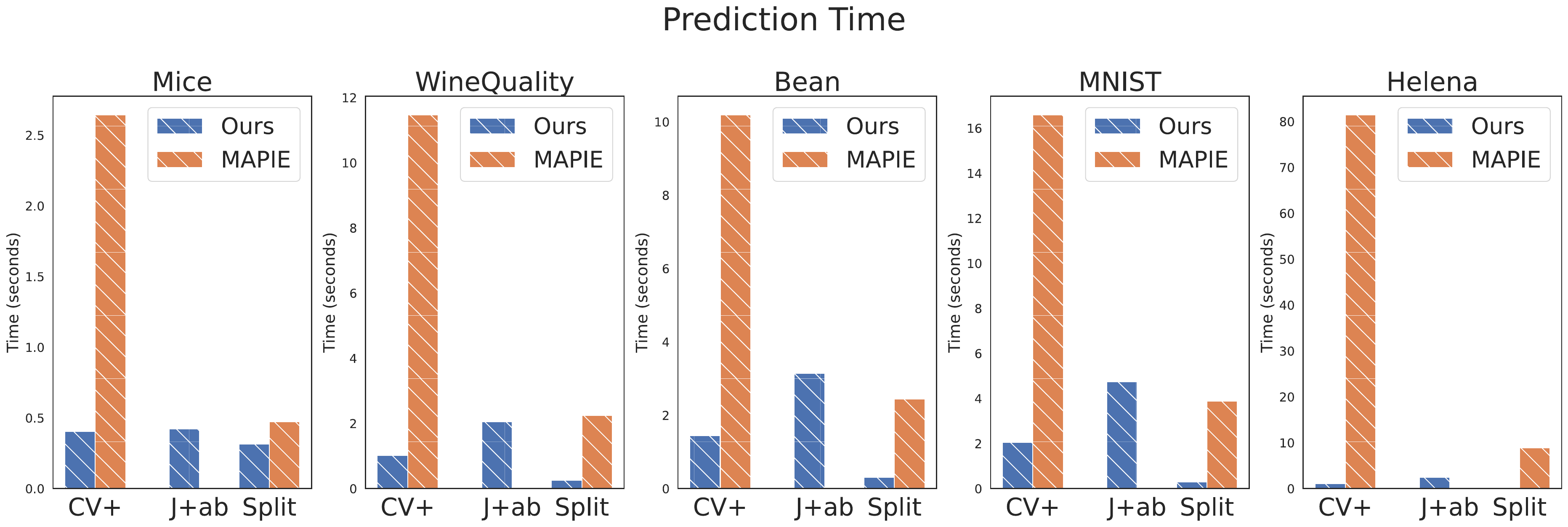}
    \caption{Training and prediction times of \covf's and \texttt{MAPIE}'s implementations of the CV+, J+ab and split conformal methods with the APS score and $\alpha=0.05$. Note that \texttt{MAPIE} does not implement the J+ab method for classification.}
    \label{fig:boxplot3}
\end{figure}

Figure~\ref{fig:boxplot3} compares the run times of both packages (note that at the time of this study, \texttt{MAPIE} had not implemented J+ab with the APS score). It shows that \covf \ consistently runs faster than \texttt{MAPIE} in both training and prediction, especially with large datasets like WineQuality and MNIST. The improvements are even more pronounced on datasets with many classes. Specifically, on the Helena dataset, \texttt{coverforest} training time is 14 times faster for the CV+ method, 8 times faster for the split method; and the testing time is 72 times faster for the CV+ method, and 30 times faster for the split method. These results indicate the performance advantage to our optimized \texttt{Numpy} code during the training phase, and the optimized \texttt{Cython} code during the prediction phase.

\subsection{Regression experiment}

\begin{table}[t]
\footnotesize
\centering

\begin{tabular}{@{}rrrrrrr@{}}
\toprule
\multicolumn{1}{l}{} &  &  & \multicolumn{2}{c}{\textbf{Experiment 1 \& 2}} & \multicolumn{2}{c}{\textbf{Experiment 3}} \\
\cmidrule(lr){4-5} \cmidrule(lr){6-7}
& \textbf{Ranges} & \textbf{Features} & \textbf{Train} & \textbf{Test} & \textbf{Train} & \textbf{Test} \\
\midrule
Housing \cite{Housing}         & [0.15, 5.0]        & 8       & 200     & 1000   & 16512 & 4128     \\
Concrete \cite{Concrete}       & [2.33, 82.6]       & 8        & 200     & 206   &   824   & 206     \\
Bike \cite{Bike}           & [1, 638]           & 12       & 200   & 1000   &  13903 & 3476    \\
Crime \cite{Crime2002}          & [0.0, 1.0]        & 122      & 200    & 399  &  1595   & 399     \\  \bottomrule
\end{tabular}
\caption{The datasets used in our experiment.}
\label{tab:regdata}
\end{table}

We now evaluate \texttt{CoverForestRegressor} for interval predictions on four regression datasets. The characteristics of the benchmark datasets are detailed in Table~\ref{tab:regdata}.

For CV+, we split the training data into 10 folds, and each random forest has 30 trees. For J+ab and split conformal, the random forest has 100 trees. We randomly sample 20\% of each dataset for the test set and 200 instances of the remaining data for the training set. For each dataset, we run each method for 50 repetitions and calculate the coverage and average interval length. The numbers of training and test instances in this experiment are indicated in Table~\ref{tab:regdata}. We fix $\alpha=0.05$. For each method, we perform training and prediction for 30 repetitions and compare the average run times to those of \texttt{crepes} and \texttt{MAPIE}.

For Experiment 1 and 2, we also perform hypothesis testing to verify that the empirical coverage probabilities exceed the nominal coverage levels, and to compare the interval lengths between any pair of the methods. The results and discussion of the tests can be found in \ref{sec:tests}.

\subsection*{Experiment 1 \& 2. Empirical coverage and average interval length}

\begin{figure}
    \centering
    \includegraphics[width=\textwidth]{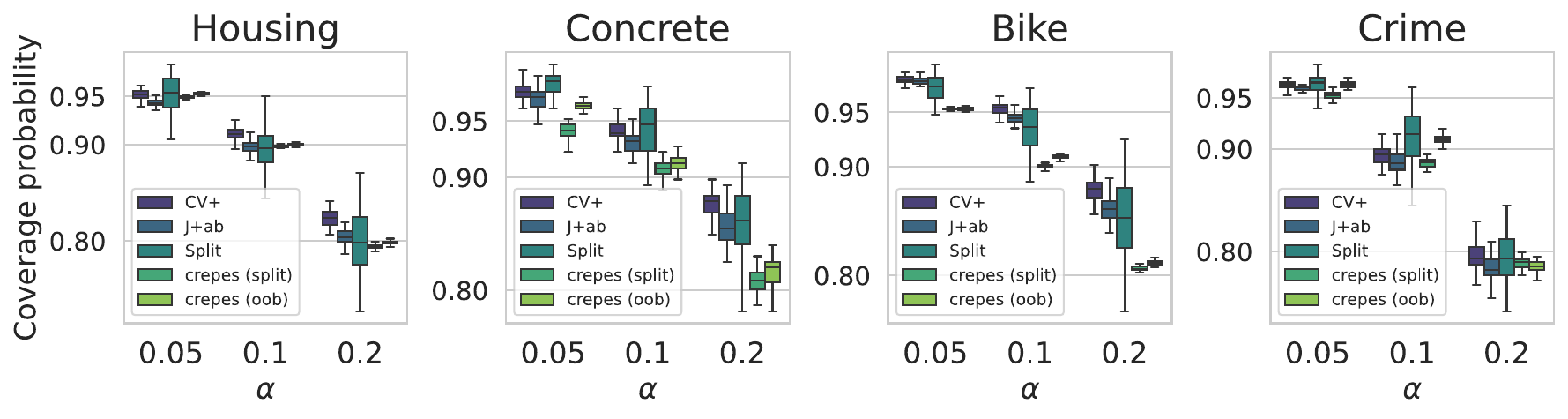}
    \includegraphics[width=\textwidth]{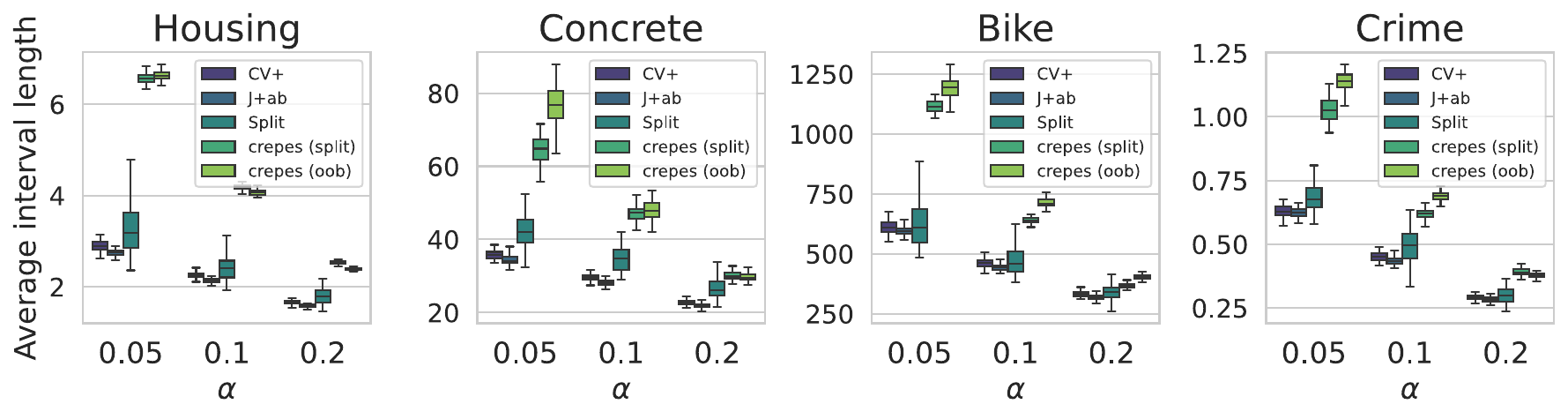}
    \caption{Coverage and average length of \texttt{CoverForestRegressor}'s interval predictions on the test set, repeated 50 times for each of the CV+, J+ab and split conformal methods with the residuals as the conformity scores. Results of \texttt{crepes}'s J+ab method are also included.}
    \label{fig:boxplot4}
\end{figure}

Figure \ref{fig:boxplot4} shows the coverages and average interval lengths of the three methods. Across all datasets, both CV+ and J+ab produce prediction intervals with small variances compared to the split conformal, with  J+ab consistently producing intervals of smaller average lengths and smaller coverages than CV+. Even though both \texttt{crepes}'s methods provide coverages that achieve the nominal levels, their interval lengths are substantially longer than those from the other methods.

\subsection*{Experiment 3. Run time comparison with \texttt{crepes} and \texttt{MAPIE}}

\begin{figure}[t]
    \centering
    \includegraphics[width=\textwidth]{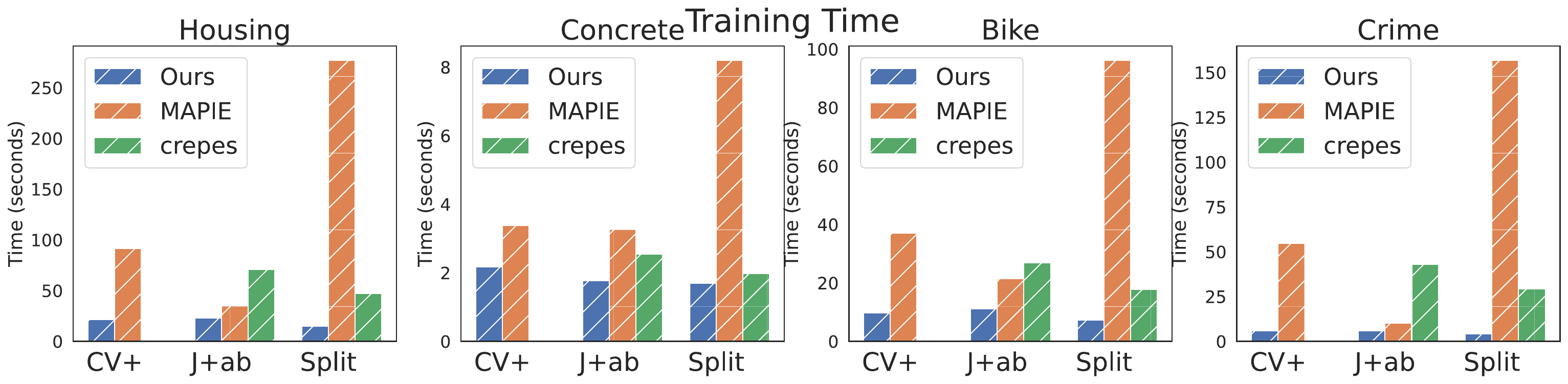}
    \includegraphics[width=\textwidth]{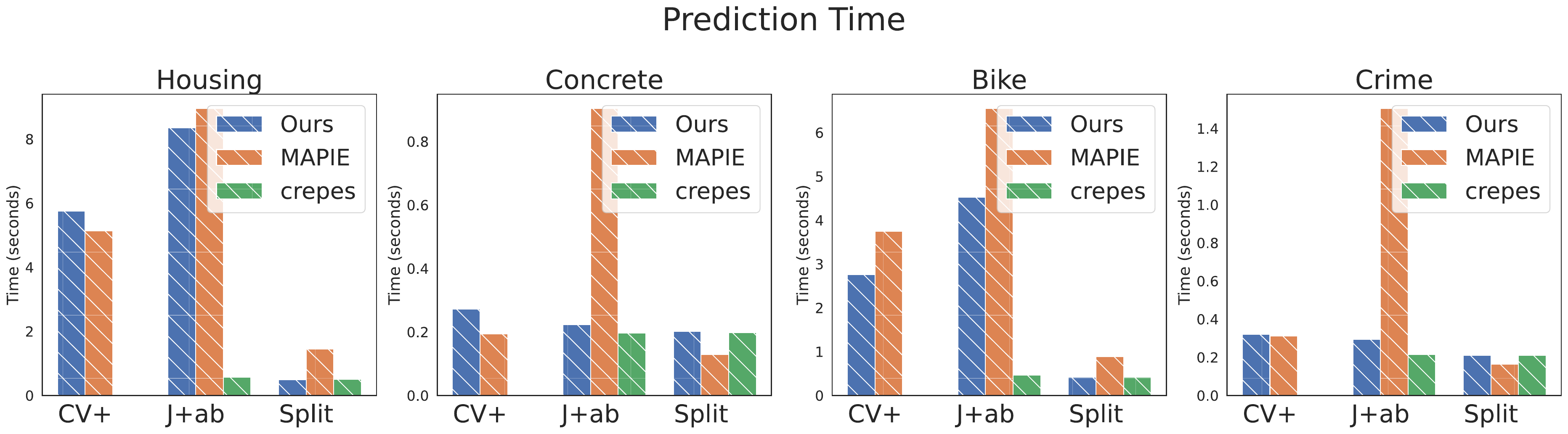}
    \caption{Training and prediction times of \covf, \texttt{crepes} and \texttt{MAPIE}'s implementations of the CV+, J+ab and split conformal methods with the residuals as the conformity scores and $\alpha=0.05$. Note that \texttt{crepes} does not implement the CV+ method.}
    \label{fig:boxplot5}
\end{figure}

Figure \ref{fig:boxplot5} presents a comparison of the run times for \texttt{coverforest}, \texttt{crepes}, and \texttt{MAPIE} on conformal regression tasks. The results clearly show that \texttt{coverforest} is substantially faster than the other packages during the training phase across all evaluated methods.

Regarding prediction time, \texttt{crepes} is generally the fastest package, noticeably with the bootstrap (J+ab) method; this is due to the implementation of the Mondrian conformal method, which only requires a single prediction for each test point. In contrast, the methods implemented in our package and in \texttt{MAPIE} necessitate making multiple predictions for each test point (more precisely, for each training point, its out-of-bag trees are used to make predictions for the test point). Nonetheless, it is important to note that, unlike in classification, prediction in conformal regression takes substantially less time than training.

\section{Conclusion}
In this work, we introduce \covf, a lightweight package that bridges the gap between \skl's optimized random forest implementation and modern conformal prediction techniques. By leveraging both cross-validation+ and Jackknife+-after-bootstrap approaches, \covf \  enables uncertainty quantification through distribution-free prediction intervals and prediction sets.

Our empirical evaluation demonstrates that \covf \  not only achieves the desired coverage guarantees but is also faster than existing implementation of conformal predictions with random forests. Given random forest's established effectiveness on tabular data, \covf \  provides a practical solution for deployment of conformal prediction in real-world applications. Future work could explore extensions to other ensemble methods and conformal prediction techniques while maintaining the package's emphasis on simplicity and efficiency.

Finally, we acknowledge a current limitation of our package regarding computational scalability. The implementation is highly optimized for multi-core CPU execution via \texttt{Cython}, but it does not currently support GPU acceleration or distributed computing. A significant avenue for future work is to extend the framework to leverage GPU-native libraries, such as \texttt{cuML} \cite{Raschka2020}, or distributed platforms like \texttt{Dask} \cite{Dask2025} and \texttt{Ray} \cite{Moritz2018} While this would require a substantial change to the codebase, it would greatly enhance the applicability of our methods to large-scale datasets that exceed the capabilities of a single machine.

\section*{Acknowledgment}

This work was supported by Chiang Mai University, Thailand.

\appendix
\section{Significant testing of the experiments' results}\label{sec:tests}

To verify the coverage guarantees and compare the methods in terms of the average set sizes, we perform two statistical tests on the results in Section \ref{sec:experiments}:
\begin{enumerate}
    \item One-sided $t$-test to verify that the empirical coverage probability for each combination of method, dataset and $\alpha$ exceeds the target of $1-\alpha$. 
    \item Two-sample $t$-test to compare the average set sizes/average interval lengths for each pair of methods in each dataset.
\end{enumerate}
For the one-sided test, we also provide $p$-values in hypotheses. For both tests, we use the star symbols: ``*'', ``**'' and ``***'' to indicate that the test's conclusion is significant at level $0.05, 0.01$ and $0.001$, respectively. For the two-sample $t$-test, the star symbol indicates the smallest significance level that holds for all values of nominal coverage $\alpha$. 

\subsection{Classification}\label{sec:testclf}

The results of the one-sided tests for the five classification datasets are shown in Table~\ref{tab:one_sided_ttest_dataset_11}, \ref{tab:one_sided_ttest_dataset_12}, \ref{tab:one_sided_ttest_dataset_13}, \ref{tab:one_sided_ttest_dataset_14} and \ref{tab:one_sided_ttest_dataset_15}. Those of the two-sample tests are shown in Table~\ref{tab:summary_size_ttest_dataset_11}, \ref{tab:summary_size_ttest_dataset_12}, \ref{tab:summary_size_ttest_dataset_13}, \ref{tab:summary_size_ttest_dataset_14} and \ref{tab:summary_size_ttest_dataset_15}. Here are some observations from these results:
\begin{itemize}
\item While the split and \texttt{crepes}' methods often produce smaller sets on average than the CV+ and J+ab methods, they frequently fail to achieve the target coverage levels, which defeats the purpose of conformal prediction. 
\item For the CV+ and J+ab methods, using RAPS scores as opposed to APS scores leads to a significant reduction in prediction set sizes, particularly for the MNIST and Helena datasets which have large numbers of classes.
\item In some cases, J+ab produces smaller set on average than CV+, but not the other way around.
\end{itemize}

\begin{table}\small\centering
\caption{One-Sided $t$-test for Coverage > $1-\alpha$ for the Mice dataset}
\label{tab:one_sided_ttest_dataset_11}
\begin{tabular}{lccc}
\toprule
 & $\alpha=0.05$ & $\alpha=0.1$ & $\alpha=0.2$ \\
Model &  &  &  \\
\midrule
CV+ (APS) & 0.970 (0.000)*** & 0.925 (0.000)*** & 0.832 (0.000)*** \\
CV+ (RAPS) & 0.970 (0.000)*** & 0.925 (0.000)*** & 0.832 (0.000)*** \\
J+ab (APS) & 0.961 (0.000)*** & 0.920 (0.000)*** & 0.826 (0.000)*** \\
J+ab (RAPS) & 0.961 (0.000)*** & 0.920 (0.000)*** & 0.825 (0.000)*** \\
Split (APS) & 0.953 (0.204) & 0.907 (0.081) & 0.814 (0.021)* \\
Split (RAPS) & 0.950 (0.532) & 0.907 (0.076) & 0.815 (0.020)* \\
\texttt{crepes} (split) & 0.970 (0.000)*** & 0.898 (0.669) & 0.737 (1.000) \\
\texttt{crepes} (oob) & 0.984 (0.000)*** & 0.940 (0.000)*** & 0.826 (0.000)*** \\
\bottomrule
\end{tabular}
\end{table}

\begin{table}\small\centering
\caption{One-Sided $t$-test for Coverage > $1-\alpha$ for the WineQuality dataset}
\label{tab:one_sided_ttest_dataset_12}
\begin{tabular}{lccc}
\toprule
 & $\alpha=0.05$ & $\alpha=0.1$ & $\alpha=0.2$ \\
Model &  &  &  \\
\midrule
CV+ (APS) & 0.961 (0.000)*** & 0.921 (0.000)*** & 0.848 (0.000)*** \\
CV+ (RAPS) & 0.957 (0.000)*** & 0.917 (0.000)*** & 0.842 (0.000)*** \\
J+ab (APS) & 0.951 (0.218) & 0.912 (0.000)*** & 0.830 (0.000)*** \\
J+ab (RAPS) & 0.948 (0.965) & 0.909 (0.000)*** & 0.821 (0.000)*** \\
Split (APS) & 0.909 (1.000) & 0.800 (1.000) & 0.638 (1.000) \\
Split (RAPS) & 0.909 (1.000) & 0.800 (1.000) & 0.629 (1.000) \\
\texttt{crepes} (split) & 0.956 (0.000)*** & 0.892 (0.893) & 0.621 (1.000) \\
\texttt{crepes} (oob) & 0.946 (0.998) & 0.909 (0.029)* & 0.722 (1.000) \\
\bottomrule
\end{tabular}
\end{table}

\begin{table}\small\centering
\caption{One-Sided $t$-test for Coverage > $1-\alpha$ for the Bean dataset}
\label{tab:one_sided_ttest_dataset_13}
\begin{tabular}{lccc}
\toprule
 & $\alpha=0.05$ & $\alpha=0.1$ & $\alpha=0.2$ \\
Model &  &  &  \\
\midrule
CV+ (APS) & 0.972 (0.000)*** & 0.934 (0.000)*** & 0.842 (0.000)*** \\
CV+ (RAPS) & 0.971 (0.000)*** & 0.933 (0.000)*** & 0.834 (0.000)*** \\
J+ab (APS) & 0.967 (0.000)*** & 0.929 (0.000)*** & 0.838 (0.000)*** \\
J+ab (RAPS) & 0.966 (0.000)*** & 0.927 (0.000)*** & 0.824 (0.000)*** \\
Split (APS) & 0.922 (1.000) & 0.875 (1.000) & 0.783 (1.000) \\
Split (RAPS) & 0.917 (1.000) & 0.872 (1.000) & 0.780 (1.000) \\
\texttt{crepes} (split) & 0.951 (0.256) & 0.913 (0.000)*** & 0.694 (1.000) \\
\texttt{crepes} (oob) & 0.974 (0.000)*** & 0.920 (0.000)*** & 0.831 (0.000)*** \\
\bottomrule
\end{tabular}
\end{table}

\begin{table}\small\centering
\caption{One-Sided $t$-test for Coverage > $1-\alpha$ for the MNIST dataset}
\label{tab:one_sided_ttest_dataset_14}
\begin{tabular}{lccc}
\toprule
 & $\alpha=0.05$ & $\alpha=0.1$ & $\alpha=0.2$ \\
Model &  &  &  \\
\midrule
CV+ (APS) & 0.971 (0.000)*** & 0.925 (0.000)*** & 0.824 (0.000)*** \\
CV+ (RAPS) & 0.970 (0.000)*** & 0.921 (0.000)*** & 0.803 (0.069) \\
J+ab (APS) & 0.958 (0.000)*** & 0.909 (0.000)*** & 0.808 (0.002)** \\
J+ab (RAPS) & 0.956 (0.000)*** & 0.905 (0.005)** & 0.786 (1.000) \\
Split (APS) & 0.927 (1.000) & 0.851 (1.000) & 0.750 (1.000) \\
Split (RAPS) & 0.927 (1.000) & 0.851 (1.000) & 0.743 (1.000) \\
\texttt{crepes} (split) & 0.935 (1.000) & 0.869 (1.000) & 0.684 (1.000) \\
\texttt{crepes} (oob) & 0.948 (0.823) & 0.865 (1.000) & 0.682 (1.000) \\
\bottomrule
\end{tabular}
\end{table}

\begin{table}\small\centering
\caption{One-Sided $t$-test for Coverage > $1-\alpha$ for the Helena dataset}
\label{tab:one_sided_ttest_dataset_15}
\begin{tabular}{lccc}
\toprule
 & $\alpha=0.05$ & $\alpha=0.1$ & $\alpha=0.2$ \\
Model &  &  &  \\
\midrule
CV+ (APS) & 1.000 (0.000)*** & 0.973 (0.000)*** & 0.858 (0.000)*** \\
CV+ (RAPS) & 0.997 (0.000)*** & 0.953 (0.000)*** & 0.837 (0.000)*** \\
J+ab (APS) & 0.998 (0.000)*** & 0.947 (0.000)*** & 0.848 (0.000)*** \\
J+ab (RAPS) & 0.984 (0.000)*** & 0.937 (0.000)*** & 0.832 (0.000)*** \\
Split (APS) & 0.983 (0.000)*** & 0.914 (0.000)*** & 0.812 (0.000)*** \\
Split (RAPS) & 0.983 (0.000)*** & 0.910 (0.000)*** & 0.806 (0.006)** \\
\texttt{crepes} (split) & 0.971 (0.000)*** & 0.938 (0.000)*** & 0.855 (0.000)*** \\
\texttt{crepes} (oob) & 0.988 (0.000)*** & 0.976 (0.000)*** & 0.878 (0.000)*** \\
\bottomrule
\end{tabular}
\end{table}

\begin{table}\small\centering
\caption{Pairwise $t$-test for Average Set Size for the Mice dataset}
\label{tab:summary_size_ttest_dataset_11}
\begin{tabular}{lllllllll}
\toprule
 & CV+ (A) & CV+ (R) & J+ab (A) & J+ab (R) & Spl (A) & Spl (R) & Cr (spl) & Cr (oob) \\
\midrule
CV+ (A) & --- & - & - & $\blacktriangle$ ** & $\blacktriangle$ ** & $\blacktriangle$ ** & $\blacktriangle$ *** & $\blacktriangle$ *** \\
CV+ (R) & - & --- & - & - & $\blacktriangle$ ** & $\blacktriangle$ * & $\blacktriangle$ *** & $\blacktriangle$ *** \\
J+ab (A) & - & - & --- & $\blacktriangle$ * & $\blacktriangle$ ** & $\blacktriangle$ ** & $\blacktriangle$ *** & $\blacktriangle$ *** \\
J+ab (R) & $\blacktriangledown$ ** & - & $\blacktriangledown$ * & --- & - & - & $\blacktriangle$ *** & $\blacktriangle$ *** \\
Spl (A) & $\blacktriangledown$ ** & $\blacktriangledown$ ** & $\blacktriangledown$ ** & - & --- & - & $\blacktriangle$ *** & $\blacktriangle$ *** \\
Spl (R) & $\blacktriangledown$ ** & $\blacktriangledown$ * & $\blacktriangledown$ ** & - & - & --- & $\blacktriangle$ *** & $\blacktriangle$ *** \\
Cr (spl) & $\blacktriangledown$ *** & $\blacktriangledown$ *** & $\blacktriangledown$ *** & $\blacktriangledown$ *** & $\blacktriangledown$ *** & $\blacktriangledown$ *** & --- & $\blacktriangledown$ *** \\
Cr (oob) & $\blacktriangledown$ *** & $\blacktriangledown$ *** & $\blacktriangledown$ *** & $\blacktriangledown$ *** & $\blacktriangledown$ *** & $\blacktriangledown$ *** & $\blacktriangle$ *** & --- \\
\bottomrule
\end{tabular}
\end{table}

\begin{table}\small\centering
\caption{Pairwise $t$-test for Average Set Size for the WineQuality dataset}
\label{tab:summary_size_ttest_dataset_12}
\begin{tabular}{lllllllll}
\toprule
 & CV+ (A) & CV+ (R) & J+ab (A) & J+ab (R) & Spl (A) & Spl (R) & Cr (spl) & Cr (oob) \\
\midrule
CV+ (A) & --- & $\blacktriangle$ * & $\blacktriangle$ *** & $\blacktriangle$ *** & $\blacktriangle$ *** & $\blacktriangle$ *** & - & $\blacktriangledown$ *** \\
CV+ (R) & $\blacktriangledown$ * & --- & - & $\blacktriangle$ *** & $\blacktriangle$ *** & $\blacktriangle$ *** & $\blacktriangledown$ ** & $\blacktriangledown$ *** \\
J+ab (A) & $\blacktriangledown$ *** & - & --- & $\blacktriangle$ * & $\blacktriangle$ *** & $\blacktriangle$ *** & $\blacktriangledown$ ** & $\blacktriangledown$ *** \\
J+ab (R) & $\blacktriangledown$ *** & $\blacktriangledown$ *** & $\blacktriangledown$ * & --- & $\blacktriangle$ *** & $\blacktriangle$ *** & $\blacktriangledown$ *** & $\blacktriangledown$ *** \\
Spl (A) & $\blacktriangledown$ *** & $\blacktriangledown$ *** & $\blacktriangledown$ *** & $\blacktriangledown$ *** & --- & - & $\blacktriangledown$ *** & $\blacktriangledown$ *** \\
Spl (R) & $\blacktriangledown$ *** & $\blacktriangledown$ *** & $\blacktriangledown$ *** & $\blacktriangledown$ *** & - & --- & $\blacktriangledown$ *** & $\blacktriangledown$ *** \\
Cr (spl) & - & $\blacktriangle$ ** & $\blacktriangle$ ** & $\blacktriangle$ *** & $\blacktriangle$ *** & $\blacktriangle$ *** & --- & - \\
Cr (oob) & $\blacktriangle$ *** & $\blacktriangle$ *** & $\blacktriangle$ *** & $\blacktriangle$ *** & $\blacktriangle$ *** & $\blacktriangle$ *** & - & --- \\
\bottomrule
\end{tabular}
\end{table}

\begin{table}\small\centering
\caption{Pairwise $t$-test for Average Set Size for the Bean dataset}
\label{tab:summary_size_ttest_dataset_13}
\begin{tabular}{lllllllll}
\toprule
 & CV+ (A) & CV+ (R) & J+ab (A) & J+ab (R) & Spl (A) & Spl (R) & Cr (spl) & Cr (oob) \\
\midrule
CV+ (A) & --- & - & - & $\blacktriangle$ *** & $\blacktriangle$ *** & $\blacktriangle$ *** & $\blacktriangle$ *** & $\blacktriangle$ *** \\
CV+ (R) & - & --- & - & $\blacktriangle$ * & $\blacktriangle$ *** & $\blacktriangle$ *** & $\blacktriangle$ *** & $\blacktriangle$ *** \\
J+ab (A) & - & - & --- & - & $\blacktriangle$ *** & $\blacktriangle$ *** & $\blacktriangle$ *** & $\blacktriangle$ *** \\
J+ab (R) & $\blacktriangledown$ *** & $\blacktriangledown$ * & - & --- & $\blacktriangle$ *** & $\blacktriangle$ *** & $\blacktriangle$ *** & $\blacktriangle$ *** \\
Spl (A) & $\blacktriangledown$ *** & $\blacktriangledown$ *** & $\blacktriangledown$ *** & $\blacktriangledown$ *** & --- & - & $\blacktriangle$ *** & $\blacktriangle$ * \\
Spl (R) & $\blacktriangledown$ *** & $\blacktriangledown$ *** & $\blacktriangledown$ *** & $\blacktriangledown$ *** & - & --- & $\blacktriangle$ *** & - \\
Cr (spl) & $\blacktriangledown$ *** & $\blacktriangledown$ *** & $\blacktriangledown$ *** & $\blacktriangledown$ *** & $\blacktriangledown$ *** & $\blacktriangledown$ *** & --- & - \\
Cr (oob) & $\blacktriangledown$ *** & $\blacktriangledown$ *** & $\blacktriangledown$ *** & $\blacktriangledown$ *** & $\blacktriangledown$ * & - & - & --- \\
\bottomrule
\end{tabular}
\end{table}

\begin{table}\small\centering
\caption{Pairwise $t$-test for Average Set Size for the MNIST dataset}
\label{tab:summary_size_ttest_dataset_14}
\begin{tabular}{lllllllll}
\toprule
 & CV+ (A) & CV+ (R) & J+ab (A) & J+ab (R) & Spl (A) & Spl (R) & Cr (spl) & Cr (oob) \\
\midrule
CV+ (A) & --- & - & $\blacktriangle$ * & $\blacktriangle$ *** & - & - & $\blacktriangle$ *** & $\blacktriangle$ *** \\
CV+ (R) & - & --- & - & $\blacktriangle$ *** & - & - & $\blacktriangle$ *** & $\blacktriangle$ *** \\
J+ab (A) & $\blacktriangledown$ * & - & --- & - & - & - & $\blacktriangle$ *** & $\blacktriangle$ *** \\
J+ab (R) & $\blacktriangledown$ *** & $\blacktriangledown$ *** & - & --- & - & - & $\blacktriangle$ *** & $\blacktriangle$ *** \\
Spl (A) & - & - & - & - & --- & - & $\blacktriangle$ *** & $\blacktriangle$ *** \\
Spl (R) & - & - & - & - & - & --- & $\blacktriangle$ *** & $\blacktriangle$ *** \\
Cr (spl) & $\blacktriangledown$ *** & $\blacktriangledown$ *** & $\blacktriangledown$ *** & $\blacktriangledown$ *** & $\blacktriangledown$ *** & $\blacktriangledown$ *** & --- & - \\
Cr (oob) & $\blacktriangledown$ *** & $\blacktriangledown$ *** & $\blacktriangledown$ *** & $\blacktriangledown$ *** & $\blacktriangledown$ *** & $\blacktriangledown$ *** & - & --- \\
\bottomrule
\end{tabular}
\end{table}

\begin{table}\small\centering
\caption{Pairwise $t$-test for Average Set Size for the Helena dataset}
\label{tab:summary_size_ttest_dataset_15}
\begin{tabular}{lllllllll}
\toprule
 & CV+ (A) & CV+ (R) & J+ab (A) & J+ab (R) & Spl (A) & Spl (R) & Cr (spl) & Cr (oob) \\
\midrule
CV+ (A) & --- & $\blacktriangle$ *** & - & $\blacktriangle$ *** & - & - & - & - \\
CV+ (R) & $\blacktriangledown$ *** & --- & - & - & - & - & - & - \\
J+ab (A) & - & - & --- & $\blacktriangle$ *** & - & - & - & - \\
J+ab (R) & $\blacktriangledown$ *** & - & $\blacktriangledown$ *** & --- & $\blacktriangledown$ * & - & - & $\blacktriangledown$ *** \\
Spl (A) & - & - & - & $\blacktriangle$ * & --- & - & - & - \\
Spl (R) & - & - & - & - & - & --- & - & - \\
Cr (spl) & - & - & - & - & - & - & --- & $\blacktriangledown$ *** \\
Cr (oob) & - & - & - & $\blacktriangle$ *** & - & - & $\blacktriangle$ *** & --- \\
\bottomrule
\end{tabular}
\end{table}

\subsection{Regression}\label{sec:testreg}

The results of the one-sided tests for the four regression datasets are shown in Table~\ref{tab:one_sided_ttest_dataset_21}, \ref{tab:one_sided_ttest_dataset_22}, \ref{tab:one_sided_ttest_dataset_23} and \ref{tab:one_sided_ttest_dataset_24}. Those of the two-sample tests are shown in Table~\ref{tab:summary_size_ttest_dataset_21}, \ref{tab:summary_size_ttest_dataset_22}, \ref{tab:summary_size_ttest_dataset_23} and \ref{tab:summary_size_ttest_dataset_24}. Here are some observations from these results:
\begin{itemize}
\item On the Concrete and Bike datasets, all methods except \texttt{crepes} (split) achieve the target coverages at $0.001$ significance level. However, for the Housing dataset, only CV+ achieves the target coverage at $0.05$ significance level or less.
\item In all cases, both \texttt{crepes}' methods produce larger intervals on average than the CV+, J+ab and split methods.
\item In all cases, the J+ab method produces smaller intervals on average than the split method.
\item On all but the Crime dataset, the J+ab method produces smaller intervals on average than the CV+ method.
\end{itemize}

\begin{table}\small\centering
\caption{One-Sided $t$-test for Coverage > $1-\alpha$ for the Housing dataset}
\label{tab:one_sided_ttest_dataset_21}
\begin{tabular}{lccc}
\toprule
 & $\alpha=0.05$ & $\alpha=0.1$ & $\alpha=0.2$ \\
Model &  &  &  \\
\midrule
CV+ & 0.951 (0.039)* & 0.910 (0.000)*** & 0.823 (0.000)*** \\
J+ab & 0.944 (1.000) & 0.897 (0.996) & 0.805 (0.000)*** \\
Split & 0.951 (0.283) & 0.896 (0.867) & 0.801 (0.432) \\
\texttt{crepes} (split) & 0.949 (1.000) & 0.898 (1.000) & 0.794 (1.000) \\
\texttt{crepes} (oob) & 0.953 (0.000)*** & 0.900 (0.196) & 0.798 (1.000) \\
\bottomrule
\end{tabular}
\end{table}

\begin{table}\small\centering
\caption{One-Sided $t$-test for Coverage > $1-\alpha$ for the Concrete dataset}
\label{tab:one_sided_ttest_dataset_22}
\begin{tabular}{lccc}
\toprule
 & $\alpha=0.05$ & $\alpha=0.1$ & $\alpha=0.2$ \\
Model &  &  &  \\
\midrule
CV+ & 0.977 (0.000)*** & 0.941 (0.000)*** & 0.877 (0.000)*** \\
J+ab & 0.969 (0.000)*** & 0.932 (0.000)*** & 0.855 (0.000)*** \\
Split & 0.982 (0.000)*** & 0.944 (0.000)*** & 0.857 (0.000)*** \\
\texttt{crepes} (split) & 0.941 (1.000) & 0.907 (0.000)*** & 0.809 (0.000)*** \\
\texttt{crepes} (oob) & 0.963 (0.000)*** & 0.915 (0.000)*** & 0.817 (0.000)*** \\
\bottomrule
\end{tabular}
\end{table}

\begin{table}\small\centering
\caption{One-Sided $t$-test for Coverage > $1-\alpha$ for the Bike dataset}
\label{tab:one_sided_ttest_dataset_23}
\begin{tabular}{lccc}
\toprule
 & $\alpha=0.05$ & $\alpha=0.1$ & $\alpha=0.2$ \\
Model &  &  &  \\
\midrule
CV+ & 0.980 (0.000)*** & 0.953 (0.000)*** & 0.879 (0.000)*** \\
J+ab & 0.979 (0.000)*** & 0.944 (0.000)*** & 0.861 (0.000)*** \\
Split & 0.973 (0.000)*** & 0.935 (0.000)*** & 0.850 (0.000)*** \\
\texttt{crepes} (split) & 0.953 (0.000)*** & 0.900 (0.074) & 0.806 (0.000)*** \\
\texttt{crepes} (oob) & 0.953 (0.000)*** & 0.909 (0.000)*** & 0.811 (0.000)*** \\
\bottomrule
\end{tabular}
\end{table}

\begin{table}\small\centering
\caption{One-Sided $t$-test for Coverage > $1-\alpha$ for the Crime dataset}
\label{tab:one_sided_ttest_dataset_24}
\begin{tabular}{lccc}
\toprule
 & $\alpha=0.05$ & $\alpha=0.1$ & $\alpha=0.2$ \\
Model &  &  &  \\
\midrule
CV+ & 0.961 (0.000)*** & 0.896 (0.993) & 0.797 (0.949) \\
J+ab & 0.958 (0.000)*** & 0.888 (1.000) & 0.785 (1.000) \\
Split & 0.963 (0.000)*** & 0.909 (0.025)* & 0.792 (0.980) \\
\texttt{crepes} (split) & 0.953 (0.000)*** & 0.885 (1.000) & 0.788 (1.000) \\
\texttt{crepes} (oob) & 0.963 (0.000)*** & 0.909 (0.000)*** & 0.785 (1.000) \\
\bottomrule
\end{tabular}
\end{table}

\begin{table}\small\centering
\caption{Pairwise $t$-test for Average Interval Length for the Housing dataset}
\label{tab:summary_size_ttest_dataset_21}
\begin{tabular}{llllll}
\toprule
 & CV+ & J+ab & Split & \texttt{crepes} (split) & \texttt{crepes} (oob) \\
\midrule
CV+ & --- & $\blacktriangle$ *** & $\blacktriangledown$ *** & $\blacktriangledown$ *** & $\blacktriangledown$ *** \\
J+ab & $\blacktriangledown$ *** & --- & $\blacktriangledown$ *** & $\blacktriangledown$ *** & $\blacktriangledown$ *** \\
Split & $\blacktriangle$ *** & $\blacktriangle$ *** & --- & $\blacktriangledown$ *** & $\blacktriangledown$ *** \\
\texttt{crepes} (split) & $\blacktriangle$ *** & $\blacktriangle$ *** & $\blacktriangle$ *** & --- & - \\
\texttt{crepes} (oob) & $\blacktriangle$ *** & $\blacktriangle$ *** & $\blacktriangle$ *** & - & --- \\
\bottomrule
\end{tabular}
\end{table}

\begin{table}\small\centering
\caption{Pairwise $t$-test for Average Interval Length for the Concrete dataset}
\label{tab:summary_size_ttest_dataset_22}
\begin{tabular}{llllll}
\toprule
 & CV+ & J+ab & Split & \texttt{crepes} (split) & \texttt{crepes} (oob) \\
\midrule
CV+ & --- & $\blacktriangle$ *** & $\blacktriangledown$ *** & $\blacktriangledown$ *** & $\blacktriangledown$ *** \\
J+ab & $\blacktriangledown$ *** & --- & $\blacktriangledown$ *** & $\blacktriangledown$ *** & $\blacktriangledown$ *** \\
Split & $\blacktriangle$ *** & $\blacktriangle$ *** & --- & $\blacktriangledown$ *** & $\blacktriangledown$ *** \\
\texttt{crepes} (split) & $\blacktriangle$ *** & $\blacktriangle$ *** & $\blacktriangle$ *** & --- & - \\
\texttt{crepes} (oob) & $\blacktriangle$ *** & $\blacktriangle$ *** & $\blacktriangle$ *** & - & --- \\
\bottomrule
\end{tabular}
\end{table}

\begin{table}\small\centering
\caption{Pairwise $t$-test for Average Interval Length for the Bike dataset}
\label{tab:summary_size_ttest_dataset_23}
\begin{tabular}{llllll}
\toprule
 & CV+ & J+ab & Split & \texttt{crepes} (split) & \texttt{crepes} (oob) \\
\midrule
CV+ & --- & $\blacktriangle$ * & - & $\blacktriangledown$ *** & $\blacktriangledown$ *** \\
J+ab & $\blacktriangledown$ * & --- & $\blacktriangledown$ * & $\blacktriangledown$ *** & $\blacktriangledown$ *** \\
Split & - & $\blacktriangle$ * & --- & $\blacktriangledown$ *** & $\blacktriangledown$ *** \\
\texttt{crepes} (split) & $\blacktriangle$ *** & $\blacktriangle$ *** & $\blacktriangle$ *** & --- & $\blacktriangledown$ *** \\
\texttt{crepes} (oob) & $\blacktriangle$ *** & $\blacktriangle$ *** & $\blacktriangle$ *** & $\blacktriangle$ *** & --- \\
\bottomrule
\end{tabular}
\end{table}

\begin{table}\small\centering
\caption{Pairwise $t$-test for Average Interval Length for the Crime dataset}
\label{tab:summary_size_ttest_dataset_24}
\begin{tabular}{llllll}
\toprule
 & CV+ & J+ab & Split & \texttt{crepes} (split) & \texttt{crepes} (oob) \\
\midrule
CV+ & --- & - & - & $\blacktriangledown$ *** & $\blacktriangledown$ *** \\
J+ab & - & --- & $\blacktriangledown$ *** & $\blacktriangledown$ *** & $\blacktriangledown$ *** \\
Split & - & $\blacktriangle$ *** & --- & $\blacktriangledown$ *** & $\blacktriangledown$ *** \\
\texttt{crepes} (split) & $\blacktriangle$ *** & $\blacktriangle$ *** & $\blacktriangle$ *** & --- & - \\
\texttt{crepes} (oob) & $\blacktriangle$ *** & $\blacktriangle$ *** & $\blacktriangle$ *** & - & --- \\
\bottomrule
\end{tabular}
\end{table}

\section{Cython Implementation for Cross-Conformity Scores}\label{sec:code}

This appendix details the Cython implementations responsible for the computational speed-up of \texttt{coverforest}. The performance gains are primarily achieved through parallel processing with OpenMP and optimized \texttt{C++} memory management that avoids Python's overhead.

We present two \texttt{Cython} functions that calculate the cross-conformity scores:
\begin{enumerate}
\item \texttt{\_compute\_predictions\_split()} for the split-conformal method
\item \texttt{\_compute\_test\_giqs\_cv()} for the CV+ and J+ab methods.
\end{enumerate}

For \texttt{\_compute\_predictions\_split()} the steps to calculate the cross-conformity scores for each test sample are:
\begin{enumerate}
    \item \textbf{Sort Probabilities}: The class probabilities $\hat{\pi}$ are sorted in descending order to establish their ranks.
    \item \textbf{Compute Cumulative Sum}: The cumulative sum of the sorted probabilities is calculated.
    \item \textbf{Apply RAPS Regularization}: The regularization penalty, controlled by \texttt{lambda\_star} and \texttt{k\_star}, is added to the cumulative scores. This penalty increases for each class included beyond the \texttt{k\_star}-th class, discouraging overly large prediction sets.
    \item \textbf{Determine Set Size (L)}: The function finds the smallest $L$ such that the RAPS score for the $L$-th ranked class is greater than the threshold $\tau$.
    \item \textbf{Randomization}: If \texttt{randomized} is true, a random draw $u$ is used to decide whether to include the $L$-th class. This is crucial for achieving exact marginal coverage by tightening the prediction sets.
\end{enumerate}
The code for \texttt{\_compute\_predictions\_split()} is shown below. The most crucial part is the \texttt{with nogil , parallel()} that allows us to release the global interpreter lock (GIL) and calculate the scores in parallel:
\begin{python}
def _compute_predictions_split(const float64_t[:,::1] oob_pred,
                                 float64_t[:,::1] out,
                                 float64_t tau,
                                 int64_t k_star,
                                 float32_t lambda_star,
                                 bint randomized,
                                 bint allow_empty_sets,
                                 intp_t num_threads,
                                 intp_t random_state):
    """Construct prediction sets from calibration set's generalized inverse quantile comformity scores (giqs).
    """
    # Variable declarations
    cdef Py_ssize_t i, j
    cdef Py_ssize_t n_samples = oob_pred.shape[0]
    cdef Py_ssize_t n_classes = oob_pred.shape[1]
    cdef int32_t* indices
    cdef float64_t* sorted_scores, *cumsum
    cdef float64_t* *penalties, *penalties_cumsum
    cdef int32_t* L
    cdef float32_t* U

    # Release the GIL and start parallel computation
    with nogil, parallel(num_threads=num_threads):
        # Allocate thread-local memory
        indices = ...
        sorted_scores = ...
        cumsum = ...
        penalties_cumsum = ...
        L = ...

        # Pre-compute penalty terms for different set sizes
        for k in prange(k_star, n_classes):
            penalties_cumsum[k] = lambda_star * (k - k_star + 1)

        # Main parallel loop over all samples
        for i in prange(n_samples):
            cdef intp_t inc = i * n_classes
            # Sort prediction probabilities and store their original indices
            _argsort(indices + inc, &oob_pred[i, 0], n_classes)

            # Compute cumulative sum of sorted probabilities
            sorted_scores[inc] = oob_pred[i, indices[inc]]
            cumsum[inc] = sorted_scores[inc]
            for j in range(1, n_classes):
                sorted_scores[inc+j] = oob_pred[i, indices[inc+j]]
                cumsum[inc+j] = cumsum[inc+j-1] 
                                  + sorted_scores[inc+j]

            # Determine the size of the prediction set (L)
            L[i] = 1
            for j in range(n_classes):
                if (cumsum[inc+j] + penalties_cumsum[j]) <= tau:
                    L[i] = j + 2
            L[i] = min(L[i], n_classes)

        # Optional randomization step to potentially shrink the set size by one
        if randomized:
            for i in prange(n_samples):
                # U[i] ~ Uniform([0, 1]).
                # For sufficiently large U[i],
                L[i] = L[i] - 1

        # Construct the final prediction sets (binary matrix)
        for i in prange(n_samples):
            inc = i * n_classes
            for j in range(L[i]):
                out[i, indices[inc+j]] = 1
\end{python}

For cross-conformal methods like CV+ and J+ab, conformity scores for test samples must be computed using the out-of-bag (OOB) predictions from the ensemble. The core logic of \texttt{\_compute\_test\_giqs\_cv()} mirrors the RAPS score calculation from the split conformal case but is applied over a rank 3 array of shape \texttt{(n\_train, n\_test, n\_classes)}. For each training sample $i$, the corresponding OOB model is used to generate scores for all test samples. This process is parallelized across the training samples, leading to a significant speed-up.

\begin{python}
def _compute_test_giqs_cv(const float64_t[:,:,::1] oob_pred,
                        float64_t[:,:,::1] out,
                        int64_t k_star,
                        float32_t lambda_star,
                        bint randomized,
                        bint allow_empty_sets,
                        intp_t num_threads,
                        intp_t random_state):
    """Compute generalized inverse quantiles for CV+ and
    Jackknife+-after-Bootstrap methods.
    """
    # Variable declarations
    cdef Py_ssize_t i, j, k
    cdef Py_ssize_t n_samples = oob_pred.shape[0]
    cdef Py_ssize_t n_test = oob_pred.shape[1]
    cdef Py_ssize_t n_classes = oob_pred.shape[2]
    cdef int32_t* I
    cdef float64_t* sorted_scores, *penalty, *E
    cdef float32_t* U

    # Release the GIL for parallel computation
    with nogil, parallel(num_threads=num_threads):
        # Allocate thread-local memory
        I = ...
        sorted_scores = ...

        # Parallel loop over training samples
        for i in prange(n_samples):
            # Inner loop over test samples
            for j in range(n_test):
                # Sort probabilities for each test sample and compute cumulative sum
                cdef intp_t ij_idx = (i*n_test+j) * n_classes
                _argsort(&I[ij_idx], &oob_pred[i,j,0], n_classes)

                for k in range(n_classes):
                    sorted_scores[ij_idx+k] = oob_pred[i,j,I[ij_idx+k]]

                # Calculate the cumulative sum of probabilities
                out[i,j,0] = sorted_scores[ij_idx]
                for k in range(1, n_classes):
                    out[i,j,k] = out[i,j,k-1] 
                                  + sorted_scores[ij_idx+k]

        # Apply regularization and randomization if specified
        if randomized:
            # Allocate thread-local memory
            penalty = ...
            E = ...

            # Pre-compute penalties
            for k in prange(k_star, n_classes):
                penalty[k] = lambda_star * (k-k_star+1)

            # Parallel loop to apply randomization and penalties
            for i in prange(n_samples):
                for j in range(n_test):
                    # U[j] ~ Uniform([0,1]).
                    ij_idx = i*samples_stride + j*n_classes
                    
                    # Calculate the RAPS score for the most 
                    # probable class. If empty sets are not 
                    # allowed, U is set to 1 for the class 
                    # to be included in the prediction set.
                    if allow_empty_sets:
                        E[ij_idx] = U[j] * out[i,j,0] + penalty[0]
                    else:
                        E[ij_idx] = out[i,j,0] + penalty[0]

                    # Iterate through the remaining classes.
                    for k in range(1, n_classes):
                        # Calculate RAPS score for the k-th class:
                        E[ij_idx+k] = (U[j]*sorted_scores[ij_idx+k] 
                                       + out[i,j,k-1] + penalty[k])

                    # Reorder the computed scores from rank order 
                    # back to the original class indices
                    for k in range(n_classes):
                        out[i,j,I[ij_idx+k]] = E[ij_idx+k]

        else: # Non-randomized case
            for i in prange(n_samples):
                for j in range(n_test):
                    for k in range(k_star, n_classes):
                        out[i,j,k] += lambda_star
\end{python}

\vskip 0.2in
\bibliography{reference}{}

\begin{thebibliography}{10}

\bibitem{vovk1999}
Volodya Vovk, Alexander Gammerman, and Craig Saunders.
\newblock Machine-learning applications of algorithmic randomness.
\newblock In {\em Proceedings of the Sixteenth International Conference on
  Machine Learning}, ICML '99, pages 444--453, San Francisco, CA, USA, 1999.
  Morgan Kaufmann Publishers Inc.

\bibitem{vovk2003}
Vladimir Vovk, David Lindsay, Ilia Nouretdinov, and Alex Gammerman.
\newblock Mondrian confidence machine.
\newblock {\em Technical Report}, 2003.

\bibitem{Lei2018}
Jing Lei, Max G’Sell, Alessandro Rinaldo, Ryan~J. Tibshirani, and Larry
  Wasserman.
\newblock Distribution-free predictive inference for regression.
\newblock {\em Journal of the American Statistical Association},
  113(523):1094--1111, June 2018.

\bibitem{Vovk2022}
Vladimir Vovk, Alexander Gammerman, and Glenn Shafer.
\newblock {\em Algorithmic Learning in a Random World}.
\newblock Springer International Publishing, 2022.

\bibitem{Romano2020}
Yaniv Romano, Matteo Sesia, and Emmanuel Candes.
\newblock Classification with valid and adaptive coverage.
\newblock In H.~Larochelle, M.~Ranzato, R.~Hadsell, M.F. Balcan, and H.~Lin,
  editors, {\em Advances in Neural Information Processing Systems}, volume~33,
  pages 3581--3591. Curran Associates, Inc., 2020.

\bibitem{kim2020}
Byol Kim, Chen Xu, and Rina Barber.
\newblock Predictive inference is free with the jackknife+-after-bootstrap.
\newblock {\em Advances in Neural Information Processing Systems},
  33:4138--4149, 2020.

\bibitem{breiman2001}
Leo Breiman.
\newblock Random forests.
\newblock {\em Machine learning}, 45:5--32, 2001.

\bibitem{Barber2021}
Rina~Foygel Barber, Emmanuel~J. Cand{\`e}s, Aaditya Ramdas, and Ryan~J.
  Tibshirani.
\newblock {Predictive inference with the jackknife+}.
\newblock {\em The Annals of Statistics}, 49(1):486 -- 507, 2021.

\bibitem{Grinsztajn2022}
Leo Grinsztajn, Edouard Oyallon, and Gael Varoquaux.
\newblock Why do tree-based models still outperform deep learning on typical
  tabular data?
\newblock In {\em Thirty-sixth Conference on Neural Information Processing
  Systems Datasets and Benchmarks Track}, 2022.

\bibitem{Fei2017}
Fei Tang and Hemant Ishwaran.
\newblock Random forest missing data algorithms.
\newblock {\em Statistical Analysis and Data Mining: An ASA Data Science
  Journal}, 10(6):363--377, 2017.

\bibitem{Mitchell2011}
Lawrence Mitchell, Terence~M. Sloan, Muriel Mewissen, Peter Ghazal, Thorsten
  Forster, Michal Piotrowski, and Arthur~S. Trew.
\newblock A parallel random forest classifier for r.
\newblock In {\em Proceedings of the Second International Workshop on Emerging
  Computational Methods for the Life Sciences}, ECMLS '11, pages 1--6, New
  York, NY, USA, 2011. Association for Computing Machinery.

\bibitem{joblib}
{Joblib Development Team}.
\newblock Joblib: running python functions as pipeline jobs, 2021.

\bibitem{cython}
S.~Behnel, R.~Bradshaw, C.~Citro, L.~Dalcin, D.S. Seljebotn, and K.~Smith.
\newblock Cython: The best of both worlds.
\newblock {\em Computing in Science Engineering}, 13(2):31 --39, March 2011.

\bibitem{angelopoulos2021un}
Anastasios~Nikolas Angelopoulos, Stephen Bates, Michael Jordan, and Jitendra
  Malik.
\newblock Uncertainty sets for image classifiers using conformal prediction.
\newblock In {\em International Conference on Learning Representations}, 2021.

\bibitem{Vovk2009}
Vladimir Vovk, Ilia Nouretdinov, and Alex Gammerman.
\newblock On-line predictive linear regression.
\newblock {\em The Annals of Statistics}, 37(3):1566--1590, 2009.

\bibitem{Dmitry2010}
Dmitry Devetyarov and Ilia Nouretdinov.
\newblock Prediction with confidence based on a random forest classifier.
\newblock In Harris Papadopoulos, Andreas~S. Andreou, and Max Bramer, editors,
  {\em Artificial Intelligence Applications and Innovations}, pages 37--44,
  Berlin, Heidelberg, 2010. Springer Berlin Heidelberg.

\bibitem{Johansson2014}
Ulf Johansson, Henrik Bostr\"{o}m, Tuve L\"{o}fstr\"{o}m, and Henrik Linusson.
\newblock Regression conformal prediction with random forests.
\newblock {\em Machine Learning}, 97(1-2):155--176, July 2014.

\bibitem{Zhang2020}
Dan~Nettleton Haozhe~Zhang, Joshua~Zimmerman and Daniel~J. Nordman.
\newblock Random forest prediction intervals.
\newblock {\em The American Statistician}, 74(4):392--406, 2020.

\bibitem{Bostrom2017}
Henrik Bostr\"{o}m, Henrik Linusson, Tuve L\"{o}fstr\"{o}m, and Ulf Johansson.
\newblock Accelerating difficulty estimation for conformal regression forests.
\newblock {\em Annals of Mathematics and Artificial Intelligence},
  81(1-2):125--144, March 2017.

\bibitem{Lofstrom2013}
Tuve L\"{o}fstr\"{o}m, Ulf Johansson, and Henrik Bostr\"{o}m.
\newblock Effective utilization of data in inductive conformal prediction using
  ensembles of neural networks.
\newblock In {\em The 2013 International Joint Conference on Neural Networks
  (IJCNN)}, pages 1--8, 2013.

\bibitem{Linusson2020}
H.~Linusson, U.~Johansson, and H.~Boström.
\newblock Efficient conformal predictor ensembles.
\newblock {\em Neurocomputing}, 397:266--278, 2020.

\bibitem{Meinshausen06}
Nicolai Meinshausen.
\newblock Quantile regression forests.
\newblock {\em Journal of Machine Learning Research}, 7(35):983--999, 2006.

\bibitem{Johnson2024}
Reid~A. Johnson.
\newblock quantile-forest: A python package for quantile regression forests.
\newblock {\em Journal of Open Source Software}, 9(93):5976, 2024.

\bibitem{Sadinle2019}
Mauricio Sadinle, Jing Lei, and Larry Wasserman.
\newblock Least ambiguous set-valued classifiers with bounded error levels.
\newblock {\em Journal of the American Statistical Association},
  114(525):223--234, June 2018.

\bibitem{Bates2021}
Stephen Bates, Anastasios Angelopoulos, Lihua Lei, Jitendra Malik, and Michael
  Jordan.
\newblock Distribution-free, risk-controlling prediction sets.
\newblock {\em J. ACM}, 68(6), September 2021.

\bibitem{angelopoulos2021a}
Anastasios~N. Angelopoulos, Stephen Bates, Emmanuel~J. Cand{\`{e}}s, Michael~I.
  Jordan, and Lihua Lei.
\newblock Learn then test: Calibrating predictive algorithms to achieve risk
  control.
\newblock {\em CoRR}, abs/2110.01052, 2021.

\bibitem{angelopoulos2024}
Anastasios~Nikolas Angelopoulos, Stephen Bates, Adam Fisch, Lihua Lei, and Tal
  Schuster.
\newblock Conformal risk control.
\newblock In {\em The Twelfth International Conference on Learning
  Representations}, 2024.

\bibitem{linusson2023nonconformist}
Henrik Linusson, Isak Samsten, Zygmunt Zaj\k{a}c, and Mart\'in Villanueva.
\newblock nonconformist: Python implementation of the conformal prediction
  framework.
\newblock \url{https://github.com/donlnz/nonconformist}, 2021.

\bibitem{Hocevar2021}
Toma\u{z} Ho\u{c}evar, Bla\u{z} Zupan, and Jonna St\r{a}lring.
\newblock Conformal prediction with orange.
\newblock {\em Journal of Statistical Software}, 98(7):1--22, 2021.

\bibitem{bostrom2024}
Henrik Bostr{\"o}m.
\newblock Conformal prediction in python with crepes.
\newblock In {\em Proc. of the 13th Symposium on Conformal and Probabilistic
  Prediction with Applications}, pages 236--249. PMLR, 2024.

\bibitem{Cordier2023}
Thibault Cordier, Vincent Blot, Louis Lacombe, Thomas Morzadec, Arnaud
  Capitaine, and Nicolas Brunel.
\newblock {Flexible and Systematic Uncertainty Estimation with Conformal
  Prediction via the MAPIE library}.
\newblock In {\em Conformal and Probabilistic Prediction with Applications},
  2023.

\bibitem{Micedata}
Clara Higuera, Katheleen~J. Gardiner, and Krzysztof~J. Cios.
\newblock Self-organizing feature maps identify proteins critical to learning
  in a mouse model of down syndrome.
\newblock {\em PLOS ONE}, 10(6):e0129126, June 2015.

\bibitem{winedata}
Paulo Cortez, António Cerdeira, Fernando Almeida, Telmo Matos, and José Reis.
\newblock Modeling wine preferences by data mining from physicochemical
  properties.
\newblock {\em Decision Support Systems}, 47(4):547--553, 2009.
\newblock Smart Business Networks: Concepts and Empirical Evidence.

\bibitem{beandata}
Murat Koklu and Ilker~Ali Ozkan.
\newblock Multiclass classification of dry beans using computer vision and
  machine learning techniques.
\newblock {\em Computers and Electronics in Agriculture}, 174:105507, 2020.

\bibitem{Automl2019}
Isabelle Guyon, Lisheng Sun-Hosoya, Marc Boull\'e, Hugo~Jair Escalante, Sergio
  Escalera, Zhengying Liu, Damir Jajetic, Bisakha Ray, Mehreen Saeed, Mich\'ele
  Sebag, Alexander Statnikov, WeiWei Tu, and Evelyne Viegas.
\newblock Analysis of the automl challenge series 2015-2018.
\newblock In {\em AutoML}, Springer series on Challenges in Machine Learning,
  2019.

\bibitem{MNISTdata}
Y.~Lecun, L.~Bottou, Y.~Bengio, and P.~Haffner.
\newblock Gradient-based learning applied to document recognition.
\newblock {\em Proceedings of the IEEE}, 86(11):2278--2324, 1998.

\bibitem{bostrom2022}
Henrik Bostr{\"o}m.
\newblock crepes: a python package for generating conformal regressors and
  predictive systems.
\newblock In {\em Proc. of the 11th Symposium on Conformal and Probabilistic
  Prediction with Applications}, pages 24--41. PMLR, 2022.

\bibitem{Housing}
R.~{Kelley Pace} and Ronald Barry.
\newblock Sparse spatial autoregressions.
\newblock {\em Statistics \& Probability Letters}, 33(3):291--297, 1997.

\bibitem{Concrete}
I.-C. Yeh.
\newblock Modeling of strength of high-performance concrete using artificial
  neural networks.
\newblock {\em Cement and Concrete Research}, 28(12):1797--1808, 1998.

\bibitem{Bike}
Hadi Fanaee-T and Joao Gama.
\newblock Event labeling combining ensemble detectors and background knowledge.
\newblock {\em Progress in Artificial Intelligence}, 2(2-3):113--127, November
  2013.

\bibitem{Crime2002}
Michael Redmond and Alok Baveja.
\newblock A data-driven software tool for enabling cooperative information
  sharing among police departments.
\newblock {\em European Journal of Operational Research}, 141(3):660--678,
  2002.

\bibitem{Raschka2020}
Sebastian Raschka, Joshua Patterson, and Corey Nolet.
\newblock Machine learning in python: Main developments and technology trends
  in data science, machine learning, and artificial intelligence.
\newblock {\em arXiv preprint arXiv:2002.04803}, 2020.

\bibitem{Dask2025}
{Dask Development Team}.
\newblock {\em Dask: Library for dynamic task scheduling}, 2016.

\bibitem{Moritz2018}
Philipp Moritz, Robert Nishihara, Stephanie Wang, Alexey Tumanov, Richard Liaw,
  Eric Liang, Melih Elibol, Zongheng Yang, William Paul, Michael~I. Jordan, and
  Ion Stoica.
\newblock Ray: a distributed framework for emerging ai applications.
\newblock In {\em Proceedings of the 13th USENIX Conference on Operating
  Systems Design and Implementation}, OSDI'18, pages 561--577, USA, 2018.
  USENIX Association.

\end{thebibliography}
\end{document}